\documentclass[10pt,twocolumn,letterpaper]{article}

\usepackage{cvpr}              

\usepackage[dvipsnames]{xcolor}
\usepackage{amsmath,amssymb,amsfonts}
\usepackage{multirow,multicol}
\usepackage{natbib}
\usepackage{algorithm}%
\usepackage{algorithmicx}%
\usepackage{graphicx}
\usepackage{textcomp}
\usepackage{xspace}
\usepackage{url}
\usepackage{dsfont}
\usepackage{booktabs}
\usepackage{balance}
\usepackage{soul}
\usepackage{pifont}
\usepackage{colortbl}
\usepackage{overpic}
\usepackage{enumitem}
\usepackage{nccmath}
\usepackage{bm}
\usepackage{algpseudocode}%
\usepackage{xspace}
\usepackage{todonotes}
\usepackage[accsupp]{axessibility} 

\usepackage[dvipsnames]{xcolor}

\definecolor{cvprblue}{rgb}{0.21,0.49,0.74}
\usepackage[pagebackref,breaklinks,colorlinks,citecolor=cvprblue]{hyperref}

\def\paperID{959} 
\def\confName{CVPR}
\def\confYear{2024}

\title{Geometrically-driven Aggregation for Zero-shot 3D Point Cloud Understanding}
\author{
Guofeng Mei \quad Luigi Riz \quad Yiming Wang \quad Fabio Poiesi\\
Fondazione Bruno Kessler, Via Sommarive, 18, 38123 Trento, Italy  \\
{\tt\small \{gmei,luriz, ywang, poiesi\}@fbk.eu}
}
\newcommand{\ourmethod}{GeoZe\xspace}
\newcommand{\vlmfeat}{VLM representation\xspace}
\newcommand{\vlmfeats}{VLM representations\xspace}
\newcommand{\geofeat}{geometric representation\xspace}
\newcommand{\geofeats}{geometric representations\xspace}

\newcommand{\knn}{$k$NN\xspace}

\newcommand{\relativeimpP}[1]{\small{\textcolor{Green}{+#1}}}
\newcommand{\relativeimpN}[1]{\small{\textcolor{BrickRed}{-#1}}}
\newcommand{\relativeimpZ}[1]{\small{0.0}}

\newcommand{\reported}[1]{\textcolor{gray}{#1}}

\begin{document}
\maketitle

\begin{abstract}
Zero-shot 3D point cloud understanding can be achieved via 2D Vision-Language Models (VLMs).
Existing strategies directly map \vlmfeats from 2D pixels of rendered or captured views to 3D points, overlooking the inherent and expressible point cloud geometric structure.
Geometrically similar or close regions can be exploited for bolstering point cloud understanding as they are likely to share semantic information.
To this end, we introduce the first training-free aggregation technique that leverages the point cloud's 3D geometric structure to improve the quality of the transferred \vlmfeats.
Our approach operates iteratively, performing local-to-global aggregation based on geometric and semantic point-level reasoning.
We benchmark our approach on three downstream tasks, including classification, part segmentation, and semantic segmentation, with a variety of datasets representing both synthetic/real-world, and indoor/outdoor scenarios.
Our approach achieves new state-of-the-art results in all benchmarks.
Code and dataset are available at \url{https://luigiriz.github.io/geoze-website/}.
\end{abstract}

\section{Introduction}\label{sec:intro}

\begin{figure}[t]
    \centering
    \includegraphics[width=0.99\columnwidth]{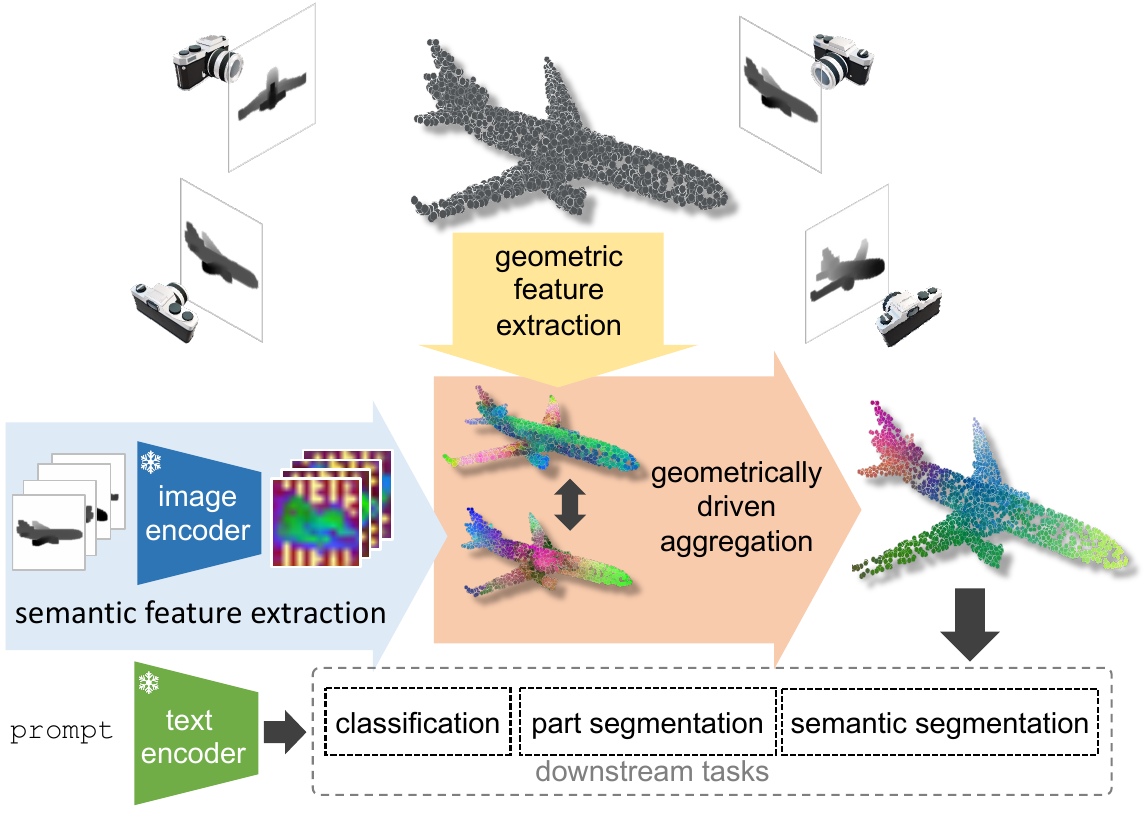}
    \caption{Given a set of dense per-pixel \vlmfeats (e.g.~CLIP~\cite{radford2021learning}) extracted from different viewpoint images of a point cloud, our approach 
    is the first geometrically-driven aggregation technique to effectively transfer these image representations to 3D points.   
    We use geometric features (e.g.~FPFH~\cite{rusu2009fast}, normals) extracted from the point cloud to denoise the \vlmfeats through an iterative process. 
    This process begins by aggregating information locally and then extends to operate globally, thereby facilitating improvements across a variety of downstream tasks.}
    \label{fig:teaser}
\end{figure}

Zero-shot point cloud understanding aims to develop deep learning models capable of performing recognition tasks, such as shape classification or segmentation, on data unobserved at training time~\cite{Peng2023, Jatavallabhula2023, zhu2023pointclip}.
Especially, vision-language models (VLMs), such as CLIP~\cite{radford2021learning}, trained on internet-scale image-text pairs, can be effectively employed for zero-shot visual recognition tasks \cite{li2023blip}.
However, training a 3D deep model using CLIP's dataset size and procedure is currently infeasible since point cloud-text pair data of the same scale as that used for CLIP is not yet available.

Transferring \vlmfeats from images to point clouds has proven to be the most effective strategy so far for achieving zero-shot point cloud understanding~\cite{ulip, zhang2022pointclip, zhu2023pointclip, Jatavallabhula2023}.
For example, PointCLIP~\cite{zhang2022pointclip} pioneers a training-free approach to directly transfer and combine per-pixel \vlmfeats of multiple rendered views of a point cloud for zero-shot classification and part segmentation.
Its successor PointCLIPv2~\cite{zhu2023pointclip} enhances 3D understanding by refining view rendering quality.
ConceptFusion~\cite{Jatavallabhula2023} fuses per-object \vlmfeats into per-pixel representations, transferring them to 3D points without requiring training. Both PointCLIPv2 and ConceptFusion have been tested in downstream tasks involving 3D scene understanding.
However, 3D points remain isolated when transferring features from 2D domain to 3D domain, lacking any bridge for intra-information flow. 
Additionally, current fusion operations often disregard the global context and spatial geometric structures.
We argue that the geometric structure of a point cloud contains valuable information that could be used to enhance the quality of the transferred \vlmfeats.
In practical terms, we can rely on the key assumption: \vlmfeats should exhibit local smoothness and global consistency when their geometric structures are similar.
To the best of our knowledge, no existing training-free method has harnessed the geometric information when transferring \vlmfeats from pixels to 3D points.

In this paper, we introduce the first \textbf{geo}metrically-driven training-free approach to aggregate point-level \vlmfeats for \textbf{ze}ro-shot point cloud understanding (\ourmethod). 
Unlike previous methods that employ naive pooling operations to transfer and aggregate \vlmfeats from images to 3D points~\cite{zhu2023pointclip,Jatavallabhula2023,Peng2023}, \ourmethod harnesses both local and global structural information to enable geometric consistency of \vlmfeats across the point cloud.
\ourmethod leverages superpoints to aggregate local information from neighboring points and facilitates a global exchange among superpoints with similar geometric structures, promoting accuracy and computational efficiency for downstream tasks.
We utilize \geofeats (3D descriptors~\cite{rusu2009fast}) to identify the structures sharing similar \vlmfeats. 
Our approach employs a weighted linear combination for aggregation, where the weights are calculated by jointly considering point cloud \vlmfeats, \geofeats, and coordinates.
A critical aspect of our method is maintaining the original alignment of visual representations in the representation space to ensure compatibility with their corresponding language representations. 
To achieve this, we introduce the concept of \vlmfeat anchors.
These anchors serve to correct potential offsets that may arise during the aggregation process, thereby preserving the integrity of the original representations.
We evaluate \ourmethod on three zero-shot downstream tasks (shape classification, part segmentation, and semantic segmentation) across five datasets (ModelNet40~\cite{wu20153d}, ObjectScanNN~\cite{uy2019revisiting}, ShapeNetPart~\cite{yi2016scalable}, ScanNet~\cite{dai2017scannet}, and nuScenes~\cite{caesar2020nuscenes}).
We use various 3D data benchmarks, including synthetic/real-world datasets, indoor/outdoor settings, and LiDAR/RGBD acquisition sensors.
\ourmethod consistently outperforms the considered baseline methods by a significant margin in a total of nine experiments.
In summary, our contributions are:
\begin{itemize}[noitemsep,nolistsep,leftmargin=*]
    \item We introduce \ourmethod, the first geometrically-driven aggregation approach for zero-shot point cloud understanding, leveraging VLMs.
    \item We show \ourmethod's versatility, being training-free and adaptable to various architectures for different downstream tasks.
    \item We establish new state-of-the-art benchmarks in zero-shot downstream tasks, including object classification, part segmentation,  and semantic segmentation.
\end{itemize}



\section{Related works}\label{sec:sota}

\noindent\textbf{Zero-shot PCD understanding.}~
The advent of large-scale VLMs has prompted a surge in zero-shot point cloud understanding~\cite{cheraghian2019mitigating, cheraghian2022zero}, including shape classification~\cite{zhang2022pointclip, ulip, hegde2023clip, huang2023clip2point}, and dense semantic segmentation~\cite{zhu2023pointclip, Peng2023, zeng2023clip2, Jatavallabhula2023}.
Zero-shot point cloud understanding can be performed by training a 3D encoder tasked with aligning 3D representations with \vlmfeats projected from images~\cite{ulip, zhu2023pointclip, Peng2023, hegde2023clip, huang2023clip2point, zeng2023clip2}.
For example, ULIP~\cite{ulip} trains a 3D encoder to align 3D features within the same latent space as the multi-view visual features extracted by CLIP's visual encoder and the textual features processed by CLIP's text encoder.
CG3D~\cite{hegde2023clip} also trains a 3D encoder and introduces prompt tuning for the visual encoder. 
CLIP2Point~\cite{huang2023clip2point} re-trains CLIP's original image encoder with multi-view depth maps using a contrastive approach.
CLIP$^2$~\cite{zeng2023clip2} dissects 3D scenes into text-image-point cloud proxies and applies cross-modal pre-training to generate point-level \vlmfeats. OpenScene~\cite{Peng2023} distills multi-view pixel-level \vlmfeats into a 3D encoder to learn point-level representations.
Zero-shot point cloud understanding can also be achieved in a training-free manner~\cite{zhang2022pointclip, zhu2023pointclip, Jatavallabhula2023}.
PointCLIP~\cite{zhang2022pointclip} renders multi-view depth maps from point clouds and aggregates view-level zero-shot predictions for 2D-to-3D knowledge transfer.
PointCLIPv2~\cite{zhu2023pointclip} incorporates large language models for 3D-specific text prompts and introduces dense depth rendering with a novel 2D-3D mapping, enabling dense point-level representation and, consequently, part segmentation tasks.
ConceptFusion~\cite{Jatavallabhula2023} seeks dense point-level representation for semantic scene understanding by exploiting off-the-shelf image segmenters to combine local crops and global representations from images to extract pixel-level features.
All the aforementioned methods do not take geometrical attributes into account during the semantic feature aggregation process. In contrast, \ourmethod incorporates geometric information during the \vlmfeat aggregation phase on the point cloud to enhance zero-shot performance across various downstream tasks. 
\ourmethod does not require any learnable parameters and can be seamlessly integrated into any training-free method.

\begin{figure*}[t]
    \centering
    \includegraphics[width=0.95\linewidth]{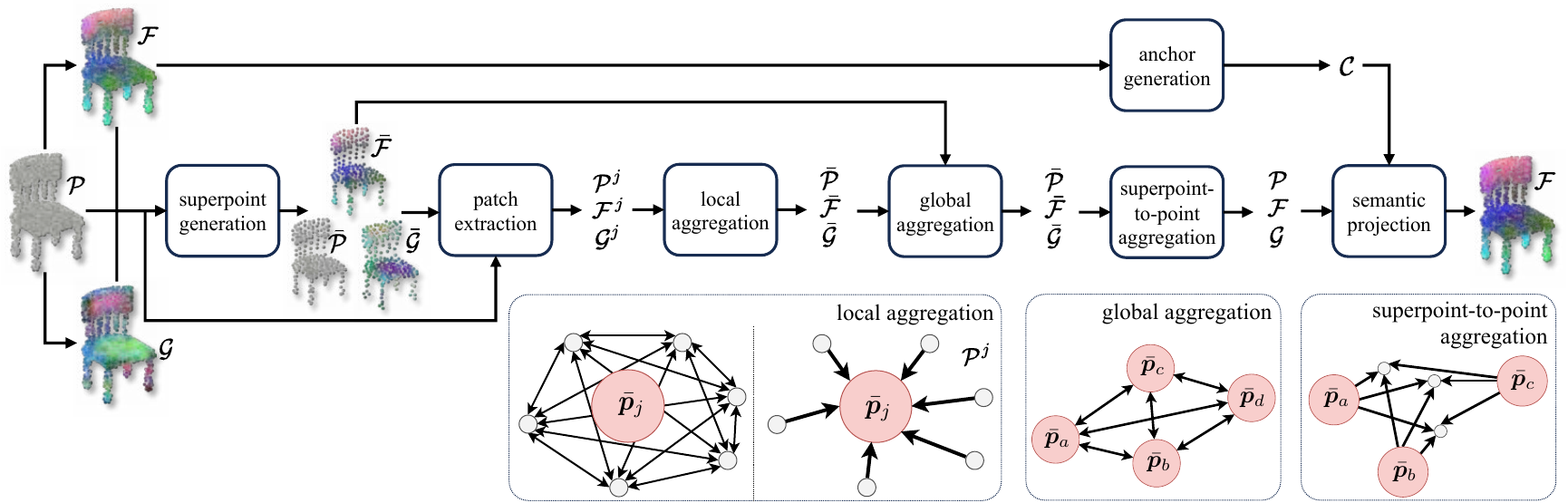}
    \vspace{-2mm}
\caption{Overview of the \ourmethod framework. \ourmethod first clusters point cloud $\bm{\mathcal{P}}$ into superpoints $\bar{\bm{\mathcal{P}}}$ along with their associated \geofeat $\bar{\bm{\mathcal{G}}}$, \vlmfeat $\bar{\bm{\mathcal{F}}}$, and anchors ${\bm{\mathcal{C}}}$. 
For each superpoint $\bar{\bm{p}_j}$, we identify its \knn within the point cloud to form a patch $\bm{\mathcal{P}}^j$ with their features $\bm{\mathcal{G}}^j$ and $\bm{\mathcal{F}}^j$.
For each patch, we perform a local feature aggregation to refine the \vlmfeats ${\bm{\mathcal{F}}}$.
The superpoints then undergo a process of global aggregation. 
A global-to-local aggregation process is applied to update the per-point features.
Lastly, we employ the VLM feature anchors to further refine per-point features, which are then ready to be utilized for downstream tasks.
}
\label{fig:block_diagram}
\vspace{-3mm}
\end{figure*}

\noindent\textbf{Point representation aggregation.}~Most methods learn to encode the geometry of point clouds via neural networks and supervised learning.
PointNet~\cite{qi2017pointnet} uses max pooling for aggregation, while PointNet++~\cite{qi2017pointnet++} hierarchically merges representations at multiple scales. 
PointCNN~\cite{li2018pointcnn} hierarchically aggregates local representations using convolution operators on a transformed feature space. 
RS-CNN~\cite{liu2019rscnn} introduces ad-hoc convolutions that consider local neighborhood relations and geometric structure for aggregation.
SPG~\cite{landrieu2018spg} models the input point clouds as hierarchical super-points graphs, in which geometrical feature aggregation is performed with graph analysis. 
DGCNN~\cite{wang2019dgcnn} learns a graph for each point cloud and applies convolution operations on its edges.
Point~Transformer~\cite{zhao2021point} introduces a transformer block to aggregate features via self-attention. 
Such operation can be computationally expensive, works like Fast~Point~Transformer~\cite{park2022fast} and Point~TransformerV2~\cite{wu2022point} mitigate this drawback.
Instead of learning semantic representations directly from 3D geometry, one can exploit multi-view rendering to learn 3D representation with additional visual inputs~\cite{wei2020view, mohammadi2021pointview, chen2020panonet3d}.
Recent zero-shot point cloud understanding methods transfer VLM representation to 3D either via learning a 3D encoder with backbones from PointNet++~\cite{qi2017pointnet++} or PointMLP~\cite{ma2022rethinking}, or by mapping dense multi-view visual representation onto the 3D point cloud without training~\cite{Jatavallabhula2023,zhu2023pointclip}. 
OpenScene~\cite{Peng2023} projects multi-view pixel-level \vlmfeats on the 3D points, and multiple \vlmfeats for a point are fused via average pooling.
ConceptFusion~\cite{Jatavallabhula2023} employs SLAM techniques to incrementally integrate pixel-level \vlmfeats into the 3D space with a momentum-based update scheme. 
Also PointCLIPv2~\cite{zhu2023pointclip} uses average pooling to fuse multi-view pixel-level \vlmfeats.
The aforementioned aggregation methods either consider 3D geometry employing learning-based methods or simply aggregate \vlmfeats without taking into account local and global geometrical structure. 
Differently, \ourmethod can effectively aggregate \vlmfeat in a geometrically-driven manner without training. 
\section{Our approach}
\subsection{Overview}
We begin by extracting dense pixel-level \vlmfeats, e.g.~with CLIP~\cite{radford2021learning}, from each viewpoint image of a point cloud.
The image can be rendered \cite{zhu2023pointclip} or be the original one used to reconstruct the point cloud \cite{Peng2023}.
These representations are then projected onto the 3D points. 
We compute point-level \geofeats, such as Fast Point Feature Histograms (FPFH)~\cite{rusu2009fast} or normal, to guide the aggregation of our \vlmfeats.
By utilizing point cloud coordinates and \geofeats, we generate superpoints along with their associated geometric and \vlmfeats. 
For each superpoint, we identify its \knn within the point cloud to form a patch.
For each patch we perform a local feature aggregation to denoise the \vlmfeats.
Then, superpoints undergo a process of global aggregation. 
We apply Mean Shift clustering on point-level \vlmfeats to establish \vlmfeat anchors. 
A global-to-local aggregation process is applied to update the point representations, based on the coordinate distance and \vlmfeats similarity between points and superpoints.
Lastly, we employ the \vlmfeat anchors to further refine point-level representations, which become then ready to be utilized for downstream tasks.
The diagram of \ourmethod is shown in Fig.~\ref{fig:block_diagram}.

\subsection{Problem formulation}

Let $\bm{\mathcal{P}} = \{\bm{p}_i \in \mathbb{R}^3 \}_{i=1}^N$ be a point cloud composed of $N$ 3D points $\bm{p}_i$.
For each $\bm{p}_i$ we compute and associate a \vlmfeat $\bm{f}_i \in\mathbb{R}^{b}$, where $b$ is the dimension of $\bm{f}_i$.
Let $\bm{\mathcal{F}} = \{\bm{f}_i \in \mathbb{R}^b \}_{i=1}^N$ be the set of \vlmfeats associated to each 3D point.
For each point $\bm{p}_i$, we compute and associate a \geofeat $\bm{g}_i \in \mathbb{R}^{d}$, where $d$ is the dimension of $\bm{g}_i$.
Let $\bm{\mathcal{G}} = \{\bm{g}_i \in \mathbb{R}^d \}_{i=1}^N$ be the set of \geofeats associated to each 3D point.
\ourmethod transforms $\bm{\mathcal{F}}$ in geometrically-coherent \vlmfeats by combining the information from $\bm{\mathcal{P}}$, $\bm{\mathcal{F}}$, and $\bm{\mathcal{G}}$ based on local and global information through an iterative aggregation process.
$\bm{\mathcal{F}}$ can be used for downstream tasks, such as classification and segmentation.

\subsection{\vlmfeat anchors and superpoints}

The cardinality of point clouds is typically large, hence we compute superpoints and perform computations on them.
A superpoint is a cluster of points featuring homogeneous geometric and \vlmfeats within a neighborhood of points.
Information can be propagated back to individual points, resulting in dense point estimations.
Intervening on \vlmfeats may lead to representations that diverge from the original feature space.
Hence, we calculate a set of \vlmfeat anchors from the superpoints, which act as a reference for determining the final point-level representations.
Steps for this computation are:
i)  initialize a set of point seeds uniformly distributed over the point cloud;
ii) compute superpoints based on both geometric and visual information;
iii) compute anchors based on aggregates of points that exhibit similar \vlmfeats.

\noindent\textbf{Seed initialization.}
We compute seeds through Farthest Point Sampling (FPS)~\cite{qi2017pointnet++}.
Specifically, we apply FPS on $\bm{\mathcal{P}}$ to sample $\bar{N}$ seeds based on point coordinates.
Let 
$\bar{\bm{\mathcal{P}}} = \{ \bar{\bm{p}}_j \in \mathbb{R}^3 \}_{j=1}^{\bar{N}}$ be the set of seeds. 
The corresponding \vlmfeats and geometric features are 
$\bar{\bm{\mathcal{F}}} = \{ \bar{\bm{f}}_j \in \mathbb{R}^b \}_{j=1}^{\bar{N}}$, and 
$\bar{\bm{\mathcal{G}}} = \{ \bar{\bm{g}}_j \in \mathbb{R}^d \}_{j=1}^{\bar{N}}$, respectively.

\noindent\textbf{Superpoint computation.}
Points proximate in the representation space are not necessarily proximate in the coordinate space. 
We aim to produce superpoints that uniformly partition the entire point cloud; thus, we formulate this as an optimal transport problem \cite{peyre2019computational,mei2023unsupervised}.
Specifically, we compute superpoints iteratively by searching for centroids in the coordinate, geometric, and \vlmfeat spaces.
This is achieved by initializing and updating their respective seeds, \(\bar{\bm{\mathcal{P}}}\), \(\bar{\bm{\mathcal{G}}}\), and \(\bar{\bm{\mathcal{F}}}\).
Firstly, for a given $j$-th seed, we select a set of $K_1$ nearest neighbors, denoted as $\mathcal{N}_j$, in the coordinate space with respect to $\bar{\bm{p}}_j$, and compute the average similarity $\bar{\bm{\mu}}=\{\bar{\bm{\mu}}_j\}$ with $\bar{\bm{\mu}}_j =\frac{1}{2K_1}\sum_{k=1}^{K_1} \left( 1.0 + \frac{\bm{\bar{f}}^k_j\bm{\bar{f}}_j}{\|\bm{\bar{f}}^k_j\|\|\bm{\bar{f}}_j\|} \right)$.
Then, we solve the following optimization problem:
\begin{equation}\label{eq:d_clu}
\vspace{-2mm}
\small
\begin{aligned}
    \min_{\bm{\gamma}} & \sum_{ij} \bm{\bm{\gamma}}_{ij}\left(\frac{1}{\sqrt{D_c}}\|\bm{p}_i-\bar{\bm{p}}_j\|_2+\frac{1}{\sqrt{D_g}}\|\bm{g}_i-\bar{\bm{g}}_j\|_2\right), \\
    & \sum_{i}\bm{\gamma}_{ij} = \bar{\bm{\mu}}_j\big/{\sum_{k}\bar{\bm{\mu}}_j},\quad \sum_{j}\bm{\gamma}_{ij} = {1}\big/{N},
    \end{aligned}
\end{equation}
where $\bm{\gamma}_{ij}$ is the probability of affiliation of $\bm{p}_i$ to $\bar{\bm{p}}_j$, and $\bm{\gamma}=\{\bm{\gamma}_{ij}\}^{N\bar{N}}_{ij}$ is the set of these probabilities.
$D_c$ and $D_g$ represent the average of the shortest distances between superpoints in the coordinate and geometric spaces, respectively. 
Eq.~\eqref{eq:d_clu} can be solved by using the Sinkhorn algorithm~\cite{peyre2019computational}.
We update the elements of $\bar{\bm{\mathcal{P}}}$, $\bar{\bm{\mathcal{G}}}$ and $\bar{\bm{\mathcal{F}}}$ as
\begin{equation*}
\small
    \bar{\bm{p}}_j {\leftarrow} \frac{\sum_{i\in\mathcal{N}_i}\bm{\gamma}_{ij}\bm{p}_i}{\sum_{i\in\mathcal{N}_i}\bm{\gamma}_{ij}}, \,
    \bar{\bm{g}}_j {\leftarrow} \frac{\sum_{i\in\mathcal{N}_i}\bm{\gamma}_{ij}\bm{g}_i}{\sum_{i\in\mathcal{N}_i}\bm{\gamma}_{ij}}, \,
    \bar{\bm{f}}_j {\leftarrow} \frac{\sum_{i\in\mathcal{N}_i}\bm{\gamma}_{ij}\bm{f}_i}{\sum_{i\in\mathcal{N}_i}\bm{\gamma}_{ij}}.
\end{equation*}
This formulation prevents superpoints that are connected in the feature space from being grouped into the same categories if they are disjoint in 3D space.
We alternately iterate $\bar{\bm{\mu}}$, $\bar{\bm{\mathcal{P}}}$, $\bar{\bm{\mathcal{G}}}$ and Eq.~\eqref{eq:d_clu} $\Gamma$ times to converge to a stable solution.

\noindent\textbf{\vlmfeat anchors.}
We model anchors as centroids of clusters characterized by homogeneous \vlmfeats and employ Mean-Shift for iterative cluster determination~\cite{kong2018recurrent}. 
We use Mean-Shift over other clustering methodologies, such as KMeans or Mixture Models, due to its ability to adaptively vary the number of clusters based on a kernel bandwidth and to be density aware. 
This approach can be particularly effective because the clusters we seek are those characterized by high-density points with distinctive representations, allowing each anchor centroid to potentially represent a semantic category.
Clusters with lower densities can instead be more likely to be representative of noisy points.
We introduce an optimized version of the Mean Shift clustering algorithm by considering both geometric and vision features.
This is implemented by initializing the  anchors with the original representations of the seeds, 
$\bm{\mathcal{C}}^0_v = \{\bm{\mathcal{C}}^{0}_{v_j} = \bar{\bm{f}}_{j} \}_{j=1}^{\bar{N}}$ and $\bm{\mathcal{C}}^0_g = \{\bm{\mathcal{C}}^{0}_{g_j} = \bar{\bm{g}}_{j} \}_{j=1}^{\bar{N}}$.
Mean-Shift is iterative and the $t$-th iteration is computed as:
\begin{equation}\label{eq:shift}
\small
   \begin{aligned}
    &\bm{\mathcal{C}}^{t{+}1}_s \leftarrow \bm{Z}\bm{K}^t{\bm{D}}^{-1},\bm{D} \leftarrow \text{diag}({\bm{K}^{t}}^{\top} \bm{1}), \\
    &\bm{K}^{t} \leftarrow \exp\left(\frac{\bm{\mathcal{F}}{\bm{\mathcal{C}}_v^t}^{\top}}{\delta_v^2\|\bm{\mathcal{F}}\|\|\bm{\mathcal{C}}_v\|}\right)
    \exp\left(\frac{\bm{\mathcal{G}}{\bm{\mathcal{C}}_g^t}^{\top}}{\delta_g^2\|\bm{\mathcal{G}}\|\|\bm{\mathcal{C}}_g\|}\right), \\
   \end{aligned}
\end{equation}
where $\delta_s (s{\in}\{v,g\})$ is the bandwidth that corresponds to the average similarity from each point to its 16-th nearest neighbor within the \vlmfeat space,  and $\|\cdot\|$ is the $L_2$ norm.
Mean-shift iterations persist until convergence, which in our experiments typically occurs around 40 iterations.
Centroids are then computed using non-maximum suppression: starting with the point of maximum density, all proximates with a similarity greater than the threshold $\delta_s / 2(s{\in}\{v,g\})$ are excluded, with the process iteratively repeated. $\delta_s (s{\in}\{v,g\})$ is the bandwidth that corresponds to the average similarity from each centroid to its 16-th nearest neighbor within the feature space. 
Points are subsequently allocated to segments, according to proximity to their closest cluster center.
We iterate this procedure about 40 times to reach the final count of centroids $L$, which serve as the label prototypes $\bm{\mathcal{C}}_v,\bm{\mathcal{C}}_g$ for subsequent analyses.


\subsection{Geometrically-driven aggregation}

Given the superpoints and \vlmfeat anchors, we perform an aggregation of \vlmfeats at point level. 
We aggregate \vlmfeats locally within the neighborhood of each superpoint and subsequently conduct a global aggregation among the superpoints. 
This information is aggregated from superpoints to individual points, assigning a \vlmfeat to each point.

\noindent\textbf{Local aggregation.}
Assuming that \vlmfeats should exhibit local smoothness, we perform local aggregation through a linear combination of \vlmfeats in the neighborhood of each superpoint.
Specifically, for the $j$-th superpoint, we select a set of $K_2$ nearest-neighbor points defined as $\bm{\mathcal{P}}^j$ for the point coordinates, $\bm{\mathcal{F}}^j$ for the point-level \vlmfeats, and \(\bm{\mathcal{G}}^j\) for the point-level geometric representations.
We compute the linear combination weights as the solution to an optimal transport problem using the Sinkhorn algorithm \cite{peyre2019computational}.
We apply the Sinkhorn algorithm based on the similarity of points in terms of their geometric and \vlmfeats.
This approach allows us to regulate the contributions from neighboring points, mitigating situations where a few highly similar points disproportionately influence the linear combination.
We define the geometric similarity as
{\small$\bm{S}_g^j = {\bm{\mathcal{G}}^j}{\bm{\mathcal{G}}^j}^{\top}\large / \sqrt{d}$}, and the \vlmfeat similarity as 
{\small$\bm{S}_v^j = \bm{\mathcal{F}}^j{\bm{\mathcal{F}}^j}^{\top} \large / \sqrt{b}$}.
Therefore, we compute the linear combination weights as {\small$\bm{\mathrm{W}}_g^j  = \mathtt{SH} \left(\bm{S}_g^j\right), \bm{\mathrm{\mathrm{W}}}_v^j  = \mathtt{SH} \left(\bm{S}_v^j\right),$}
where $\mathtt{SH}$ is the Sinkhorn function with 5 iterations.
We combine, normalize, and use these weights to aggregate the \vlmfeats
\begin{equation}
\small
\bm{\mathrm{W}}^{j} = \mathtt{SM}(\bm{\mathrm{W}}_g^j*\bm{\mathrm{\mathrm{W}}}_v^j), \quad
\bm{\mathcal{F}}^j = \frac{1}{2}\left(\bm{\mathcal{F}}^j + \bm{\mathrm{W}}^{j}\bm{\mathcal{F}}^j\right),
\end{equation}
where $\mathtt{SM}$ is the Softmax operation.
Then, we use the refined local features ($\bm{\mathcal{F}}^j$) to update the superpoint features by $\bar{\bm{\mathcal{F}}}=\bm{\gamma}\bar{\bm{\mathcal{F}}}$.
$\bm{\gamma}$ is same as in Eq.~\eqref{eq:d_clu}.
Once all the superpoints are processed, the next step is the global aggregation.

\noindent\textbf{Global aggregation.}
Geometrically similar regions may belong to similar objects, hence their \vlmfeats should be similar too.
In order to aggregate \vlmfeats of objects with similar geometric features we expand our search at global level.
As for the local aggregation, we use the Sinkhorn algorithm based on the similarity of geometric and \vlmfeats.
Let $\bar{\bm{S}}^g$ be the geometric similarity computed among all superpoints, where each element is computed as 
$\bar{\bm{S}}^g_{ij} = \bar{\bm{g}}_i\bar{\bm{g}}^\top_j \large/ \sqrt{d}$.
Similarly, let $\bm{S}^v$ be the \vlmfeat similarity, where each element is computed as 
$\bar{\bm{S}}^v_{ij} = \bar{\bm{f}}_i\bar{\bm{f}}^\top_j \large/ \sqrt{b}$.
The weights $\bar{\bm{W}}$ for the global aggregation are expressed as:
\begin{equation*}
\bar{\bm{W}}^g {=} \mathtt{SH} \left(\bar{\bm{S}}^g\right),
\bar{\bm{W}}^v {=} \mathtt{SH} \left(\bar{\bm{S}}^v\right), 
\bar{\bm{W}} {=} \mathtt{SM}(\bar{\bm{W}}^g*\bar{\bm{W}}^v*\bar{\bm{M}}),
\end{equation*}
where $\bar{\bm{M}}$ is a mask matrix with their element $\bar{\bm{M}}_{ij}=1$ if $\|\bar{\bm{p}}_i-\bar{\bm{p}}_j\|_2 < D_c$, otherwise $\bar{\bm{M}}_{ij}=0$. Then, the superpoint level features are updated by {\small$\bm{\bar{\mathcal{F}}} =  \frac{1}{2}\left( \bm{\bar{\mathcal{F}}} + \bar{\bm{W}}\bm{\bar{\mathcal{F}}}\right)$}.

\noindent\textbf{Superpoint-to-point aggregation.}
We aggregate the updated superpoints' \vlmfeats together with their neighboring points.
We use the distance between the point and each superpoint to weight the contribution of the \vlmfeat to aggregate.
Specifically, given a query point $\bm{p}_i \in \bm{\mathcal{P}}$,  the weight of superpoint $\bar{\bm{p}}_j\in \bar{\bm{\mathcal{P}}}$ to query point is decided by their coordinate distance 
$\bm{S}_{ij}^c {=} \mathtt{SM}\left(D_c{-}\|{\bm{p}}_i{-}\bar{\bm{p}}_j\|_2\right)$ and feature similarity $\bm{S}_{ij}^v=\bm{f}_i\bar{\bm{f}}^\top_j \large/ \sqrt{b}$.
We opted for the tanh kernel due to its ability to sharpen the distance measure, thereby ensuring that points are primarily influenced by their closest superpoints in the coordinate space.
Let $\bm{S}^c {=} \{\bm{S}_{ij}^c\}_{i,j=1,1}^{N,\bar{N}}$ and $\bm{S}^v {=} \{\bm{S}_{ij}^v\}_{i,j=1,1}^{N,\bar{N}}$.
Then, the weight $\bm{W}$ is calculated by 
\begin{equation}
\small
\bm{W}^c = \mathtt{SH} \left({\bm{S}}^c\right),
\bm{W}^v = \mathtt{SH} \left({\bm{S}}^v\right), 
\bm{W} = \mathtt{SM}(\bm{W}^c*{\bm{W}}^v).
\end{equation}
The point-level feature is updated via $\bm{\mathcal{F}} = \frac{1}{2}\left(\bm{\mathcal{F}} + \bm{\mathrm{W}}\bar{\bm{\mathcal{F}}}\right)$ to produce our final aggregated representations.

\noindent\textbf{Anchor projection.}~Noisy representations may be incorporated during aggregation, thus causing offsets within the \vlmfeat space.
Therefore, we project \vlmfeats onto our \vlmfeat anchors as $\bm{c}_i = \arg\max_{\bm{c}_j\in \bm{\mathcal{C}}}\left({\bm{f}_i^\top\bm{\mathcal{C}}_v}(\bm{g}_i^\top\bm{\mathcal{C}}_g\right))$
This makes the aggregated \vlmfeats to be more inclined to resemble their corresponding semantic representations.

\section{Results}

\subsection{Experimental setup}

\ourmethod is implemented using PyTorch and Open3D libraries.
Experiments are executed on an NVIDIA A40 48GB GPU. 
We use FPFH~\cite{rusu2009fast} as geometric features.
Additional setting details are in the Supplementary Material.

\noindent\textbf{Prompting.}~We use the text prompting techniques proposed in the comparison methods. 
Specifically, for PointCLIPv2~\cite{zhu2023pointclip} we use the prompts obtained via ``GPT Prompting''~\cite{zhu2023pointclip}.
For OpenScene~\cite{Peng2023} and ConceptFusion~\cite{Jatavallabhula2023}, we use the ``prompt engineering'' proposed in~\cite{Peng2023}.

\subsection{Shape Classification}\label{sec:exp:cls}

\noindent\textbf{Setting.}~We evaluate \ourmethod on four datasets for shape classification tasks. Specifically, we use the synthetic dataset ModelNet40~\cite{wu20153d}, and three variants of the real-world dataset ObjectScanNN~\cite{uy2019revisiting}.
ModelNet40~\cite{wu20153d} comprises 12,311 point clouds, with 2,468 of these designated for testing, uniformly sampled from CAD mesh surfaces across 40 categories.
We report the results on the test set.
ScanObjectNN~\cite{uy2019revisiting} comprises 15 categories and 2,902 point clouds divided in: 
OBJ-ONLY (ground-truth segmented objects extracted from the scene beforehand, i.e., objects without background), 
OBJ-BG (objects attached with background data), 
and 
S-PB-T50-RS (objects perturbed with random rotation and scaling).
We report the results computed on the test set that comprises 580 point clouds.
\ourmethod processes the depth maps extracted with PointCLIPv2~\cite{zhu2023pointclip}.
Unlike PointCLIPv2 that performs average pooling on the global representations of each depth map, we transform depth maps in point clouds, apply GeoZe, and then perform max pooling on the new \vlmfeats.

\begin{figure}[t]
\centering
\subfloat[PointCLIPv2~\cite{zhu2023pointclip}]{\includegraphics[width=0.485\columnwidth]{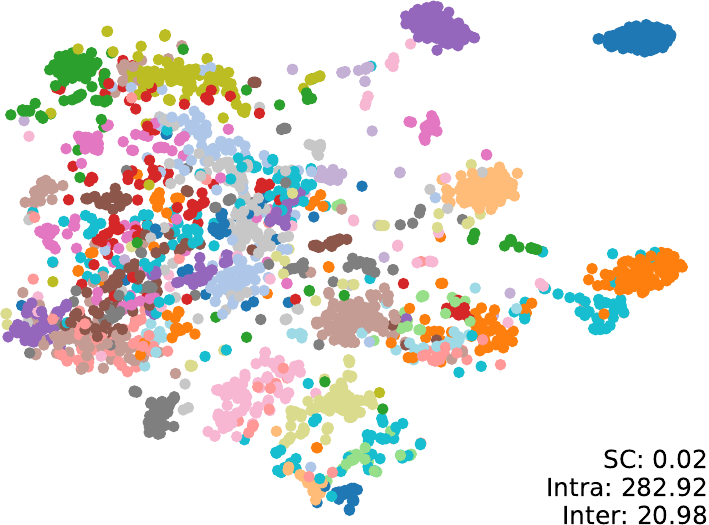}}
\hfil
\subfloat[\ourmethod]{\includegraphics[width=0.49\columnwidth]{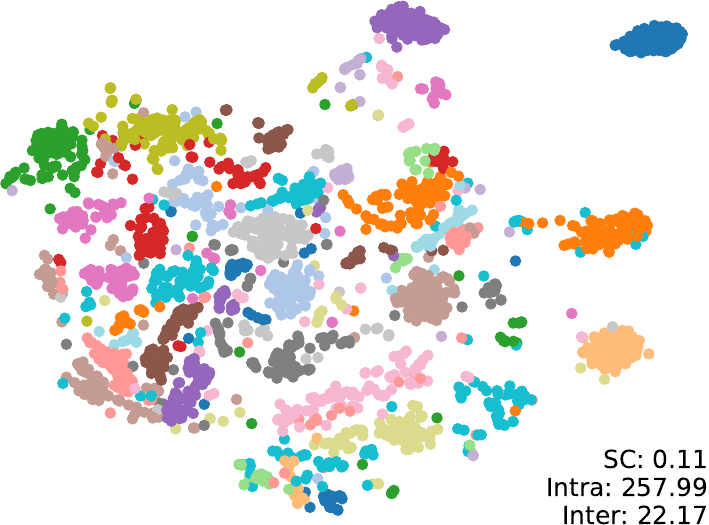}}
\vspace{-2mm}
\caption{T-SNE embeddings of (a) PointCLIPv2~\cite{zhu2023pointclip} and (b) \ourmethod on ModelNet40. 
\ourmethod produces better separated and grouped clusters for different categories, as evidenced by the superior silhouette coefficient (SC) and greater inter-cluster distance (inter), alongside a smaller intra-cluster distance (intra). 
}
\label{fig:tsne}
\vspace{-4mm}
\end{figure}

\noindent\textbf{Results.}~Tab.~\ref{tab:zs_cls} reports comparative results in terms of classification accuracy.
\ourmethod outperforms PointCLIPv2 on all datasets. 
Specifically, \ourmethod achieves $70.17\%$ on ModelNet40, outperforming PointCLIPv2 by $+5.95$. 
Moreover, \ourmethod achieves $59.34\%$, $46.00\%$, and $39.87\%$ on the three variations of ScanObjectNN.
The margin with respect to PointCLIPv2 is significant, i.e.,~$+18.03$, $+8.13$, and $+4.37$ accuracy, respectively.
In Fig.~\ref{fig:tsne}, we report a t-SNE comparison between the class representations extracted with PointCLIPv2 and \ourmethod.
We quantify t-SNE clusters with clustering metrics: Silhouette Coefficient, Inter-cluster Distance, and Intra-cluster Distance. 
The metrics underscore the efficacy of \ourmethod in more effectively separating point features from diverse categories.

\renewcommand{\arraystretch}{0.9}
\begin{table}[t]
\small
\centering
\caption{Zero-shot 3D classification results on ModelNet40~\cite{sharma2016vconv}, and ScanObjectNN~\cite{uy2019revisiting} in terms of accuracy. 
Bold font denotes best performance.
Keys: MN40: ModelNet40, SOO: S-OBJ-ONLY, SOB: S-OBJ-BG, SPB: S-PB-T50-RS. repo.: reported from the paper. repr.: reproduced by us.}
\label{tab:zs_cls}
\vspace{-3mm}
\tabcolsep 9pt
\resizebox{.85\columnwidth}{!}{%
\begin{tabular}{l | c c c c}
\toprule
Method & MN40 & SOO & SOB & SPB \\
\midrule 
\reported{PointCLIPv2 (repo.)} & \reported{64.2} & \reported{50.1} & \reported{41.2} & \reported{35.4} \\
PointCLIPv2 (repr.) & 63.3 & 41.3 & 37.9 & 35.5\\
\ourmethod & \bf70.2 & \bf59.3 & \bf46.0 & \bf39.9 \\
\midrule
$\Delta_\text{Accuracy}$ & \relativeimpP{6.8} & \relativeimpP{18.0} & \relativeimpP{8.1} & \relativeimpP{4.4} \\
\bottomrule
\end{tabular}
}
\end{table}

\renewcommand{\arraystretch}{0.9}
\begin{table*}[ht]
\centering
\caption{Zero-shot part segmentation results on ShapeNetPart~\cite{yi2016scalable} in terms of mean Intersection over Union (mIoU). 
\ourmethod outperforms PointCLIPv2's~\cite{zhu2023pointclip} on 14 out of 16 classes. 
Bold denotes best performance. Key: repo.: reported from the paper. repr.: reproduced by us.}
\label{tab:part_seg}
\vspace{-3mm}
\tabcolsep 5pt
\resizebox{\linewidth}{!}{%
\begin{tabular}{l|c| c c c c c c c c c c c c c c c c}
\toprule
& mIoU & \rotatebox{45}{Airpl.} & \rotatebox{45}{Bag} & \rotatebox{45}{Cap} & \rotatebox{45}{Car} & \rotatebox{45}{Chair} & \rotatebox{45}{Earph.} & \rotatebox{45}{Guitar} & \rotatebox{45}{Knife} & \rotatebox{45}{Lamp} & \rotatebox{45}{Laptop}  & \rotatebox{45}{Motor.} & \rotatebox{45}{Mug} & \rotatebox{45}{Pistol} & \rotatebox{45}{Rocket} & \rotatebox{45}{Skate} & \rotatebox{45}{Table}\\
\midrule
\reported{PointCLIPv2 (repo.)} & \reported{49.5} & \reported{33.5} & \reported{60.4} & \reported{52.8} & \reported{-} & \reported{51.5} & \reported{56.5} & \reported{71.5} & \reported{66.7} & \reported{-} & \reported{61.6} & \reported{-} & \reported{48.0} & \reported{-} & \reported{\textbf{49.6}} & \reported{43.9} & \reported{61.1}\\
PointCLIPv2 (repr.) & 51.8 & 33.5 & 60.4 & 52.9 & 27.2 & 51.5 & 56.5 & 71.5 & 76.7 & 44.7 & 61.5 & \bf31.5 & 48.0 & 46.1 & \bf49.6 & 43.9 & 61.1\\
\ourmethod (ours) & \bf57.4 & \bf33.6 & \bf70.2 & \bf64.7 & \bf33.6 & \bf66.1 & \bf63.7 & \bf73.6 & \bf77.9 & \bf45.4	& \bf74.0 & 30.5 & \bf63.2 & \bf48.3 & 47.3 & \bf45.6 & \bf63.8\\
\midrule
$\Delta_\text{IoU}$  & \relativeimpP{5.6} & \relativeimpP{0.1} & \relativeimpP{9.8} & \relativeimpP{11.8} & \relativeimpP{7.4} & \relativeimpP{14.6} & \relativeimpP{7.2}& \relativeimpP{2.1} & \relativeimpP{1.2} & \relativeimpP{0.7} & \relativeimpP{13.5} & \relativeimpN{1.0} & \relativeimpP{15.2} & \relativeimpP{2.5} & \relativeimpN{2.3} & \relativeimpP{1.7} & \relativeimpP{2.7}\\
\bottomrule
\end{tabular}
}
\end{table*}
\begin{figure*}[t]
\centering
\includegraphics[width=0.95\textwidth]{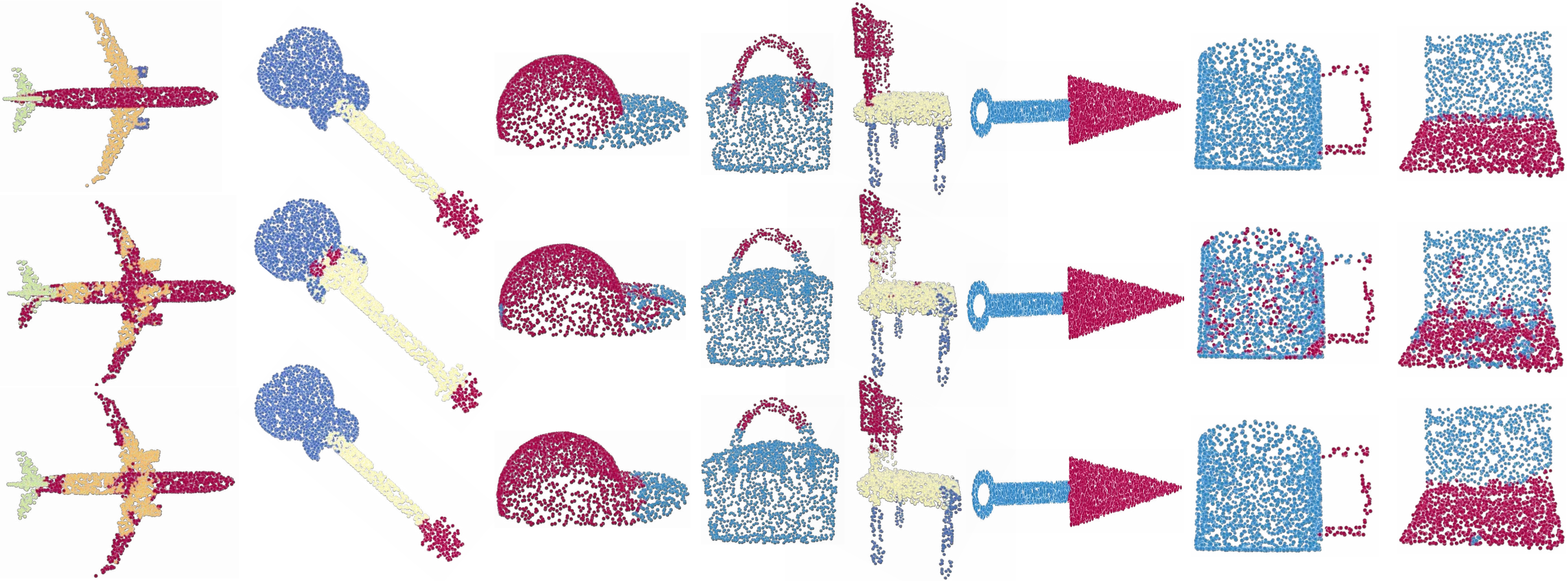}
\put(-480,140){\color{black}\footnotesize \rotatebox{90}{\textbf{GT}}}
\put(-480,70){\color{black}\footnotesize \rotatebox{90}{\textbf{PointCLIPv2}}}
\put(-480,20){\color{black}\footnotesize \rotatebox{90}{\textbf{\ourmethod}}}
\vspace{-3mm}
\caption{Zero-shot part segmentation results on ShapeNetPart \citep{yi2016scalable}.
(top row) ground-truth annotations, (middle row) PointCLIPv2 \cite{zhu2023pointclip}, and (bottom row) \ourmethod.
Parts segmented by \ourmethod are more homogeneous than those segmented by PointCLIPv2.
}
\label{fig:part_seg}
\vspace{-2mm}
\end{figure*}

\subsection{Part Segmentation} \label{sec:exp:partseg}

\noindent\textbf{Setting.}~We evaluate \ourmethod on ShapeNetPart~\cite{yi2016scalable} for the part segmentation task.
We compare \ourmethod against PointCLIPv2~\cite{zhu2023pointclip}.
ShapeNetPart includes 2,874 different objects, divided in 16 categories, and containing a total of 50 different part labels. These annotations are at point level.
Following the evaluation procedure in PointCLIPv2~\cite{zhu2023pointclip}, we randomly sample 2,048 points from each point cloud. 

\noindent\textbf{Results.} 
Tab.~\ref{tab:part_seg} reports the results in terms of the mean intersection of union (mIoU). 
For PointCLIPv2, we report the results presented in the original paper and the ones we reproduced by executing their released source code.
\ourmethod outperforms PointCLIPv2 by $+5.6$, achieving $57.4$ mIoU.
\ourmethod improves the part segmentation quality for 14 out of 16 objects types, with very large margin in the case of 
\textit{chair} ($+14.6$), 
\textit{earphone} ($+7.2$), 
\textit{laptop} ($+13.5$) and 
\textit{mug} ($+15.2$).
Fig.~\ref{fig:part_seg} reports qualitative part segmentation examples obtained with PointCLIPv2 and \ourmethod.
\ourmethod can effectively denoise features, thus producing more homogeneous segmented parts. 
There are cases, like the aeroplane, that are difficult to improve due to the high level of noise.

\renewcommand{\arraystretch}{0.9}
\begin{table*}[!h]
\centering
\caption{Zero-shot semantic segmentation results on ScanNet~\cite{dai2017scannet} in terms of mean Intersection over Union (mIoU).
\ourmethod outperforms the comparison methods (OpenScene~\cite{Peng2023} and ConceptFusion~\cite{Jatavallabhula2023}) on all datasets. 
Bold font indicates best performance.
}
\label{tab:indoor_sem_seg}
\vspace{-3mm}
\tabcolsep 3pt
\resizebox{1\linewidth}{!}{%
\begin{tabular}{l|c| c c c c c c c c c c c c c c c c c c c c }
\toprule
~ & mIoU & \rotatebox{45}{Wall} & \rotatebox{45}{Floor} & \rotatebox{45}{Cabin.} & \rotatebox{45}{Bed} & \rotatebox{45}{Chair} & \rotatebox{45}{Sofa} & \rotatebox{45}{Table} & \rotatebox{45}{Door} & \rotatebox{45}{Wind.} & \rotatebox{45}{Books.} & \rotatebox{45}{Pict.} & \rotatebox{45}{Count.} & \rotatebox{45}{Desk} & \rotatebox{45}{Curt.} & \rotatebox{45}{Refri.} & \rotatebox{45}{Sh. C.} & \rotatebox{45}{Toil.} & \rotatebox{45}{Sink} & \rotatebox{45}{Batht.} & \rotatebox{45}{Oth. F.} \\
\midrule
\multicolumn{22}{c}{OpenScene~\cite{Peng2023} (OpenSeg)}\\
\midrule
\reported{Reported} & \reported{41.4} & \reported{-} & \reported{-} & \reported{-} & \reported{-} & \reported{-} & \reported{-} & \reported{-} & \reported{-} & \reported{-} & \reported{-} & \reported{-} & \reported{-} & \reported{-} & \reported{-} & \reported{-} & \reported{-} & \reported{-} & \reported{-} & \reported{-} & \reported{-} \\
Reproduced & 43.6 & 58.4 & 72.3 & 33.5 & 66.2 & 45.4 & 48.6 & 31.7 & \bf40.1 & 50.3 & 59.8 & \bf 23.7 & 31.9 & 25.9 & \bf63.3 & 26.2 & 40.1 & 59.8 & 19.9 & 69.0 & \bf6.1 \\
\ourmethod (ours) & \bf45.6 & \bf64.2 & \bf81.0  & \bf33.7  & \bf68.8  & \bf49.0 & \bf50.6 & \bf35.6  & 39.4  & \bf53.4  & \bf62.7  & 18.9  & \bf35.1  & \bf28.3  & 61.9  & \bf28.8  & \bf42.0 & \bf63.9  & \bf20.4  & \bf75.2  & 4.0  \\
\midrule
$\Delta_\text{IoU}$ & \relativeimpP{2.0} & \relativeimpP{5.8} & \relativeimpP{8.7}  & \relativeimpP{0.6}  & \relativeimpP{2.6}  & \relativeimpP{3.6} & \relativeimpP{1.6} & \relativeimpP{3.9}  & \relativeimpN{0.7}  & \relativeimpP{3.1}  & \relativeimpP{2.9}  & \relativeimpN{4.8}  & \relativeimpP{3.2}  & \relativeimpP{2.4}  & \relativeimpN{2.4}  & \relativeimpP{2.6}  & \relativeimpP{1.9} & \relativeimpP{4.1}  & \relativeimpP{0.5} & \relativeimpP{6.2}  & \relativeimpN{2.1}  \\
\midrule
\multicolumn{22}{c}{OpenScene~\cite{Peng2023} (LSeg)}  \\
\midrule
\reported{Reported} & \reported{50.0} & \reported{-} & \reported{-} & \reported{-} & \reported{-} & \reported{-} & \reported{-} & \reported{-} & \reported{-} & \reported{-} & \reported{-} & \reported{-} & \reported{-} & \reported{-} & \reported{-} & \reported{-} & \reported{-} & \reported{-} & \reported{-} & \reported{-} & \reported{-} \\
Reproduced & 51.9 & 72.1 & 81.9 & 45.0 & 74.0 & 60.1 & 55.3 & 49.4 & 52.4 & 53.9 & 68.6 & 27.4 & 37.2 & 42.1 & 56.8 & 50.9 & 0.0 & 68.7 & \bf47.9 & 62.8 & 31.0 \\
\ourmethod (ours) & \bf55.8 & \bf75.5 & \bf88.4 & \bf47.3 & \bf78.1 & \bf67.9 & \bf58.3 & \bf54.5 & \bf53.3 & \bf59.4 & \bf73.6 & \bf26.7 & \bf44.6 & \bf46.5 & \bf57.0 & \bf56.0 & 0.0 & \bf75.5 & 47.1 & \bf71.4 & \bf33.6  \\
\midrule
$\Delta_\text{IoU}$ & \relativeimpP{3.9} & \relativeimpP{3.4} & \relativeimpP{7.5} & \relativeimpP{2.3} & \relativeimpP{4.1} & \relativeimpP{2.8} & \relativeimpP{3.0} & \relativeimpP{5.1} & \relativeimpP{0.9} & \relativeimpP{1.5} & \relativeimpP{5.0} & \relativeimpN{0.7} & \relativeimpP{7.4} & \relativeimpP{4.4} & \relativeimpP{0.2} & \relativeimpP{5.1} & \relativeimpZ{0.0} & \relativeimpP{9.6} & \relativeimpN{0.8} & \relativeimpP{8.6} & \relativeimpP{2.6}  \\
\midrule
\multicolumn{22}{c}{ConceptFusion~\cite{Jatavallabhula2023}}\\
\midrule
Reproduced & 23.4 & 17.9 & 38.6 & 22.4 & 44.7 & 31.8 & 33.8 & 24.9 & 36.5 & 20.1 & 36.7 & 8.4 & 2.6 & 9.8 & 30.7 & 13.5 & 19.1 & 13.8 & 17.3 & 39.8 & 5.8 \\
\ourmethod (ours) & \bf26.5 & \bf21.9 & \bf43.9 & \bf23.3 & \bf46.4 & \bf32.8 & \bf35.1 & \bf25.4 & \bf38.4 & \bf23.4 & \bf42.3 & 8.0 & \bf3.0 & \bf10.8 & \bf34.4 & \bf16.6 & \bf22.7 & \bf17.2 & \bf17.7 & \bf58.3 & \bf6.5 \\
\midrule
$\Delta_\text{IoU}$ & \relativeimpP{3.1} & \relativeimpP{4.0} & \relativeimpP{5.2} & \relativeimpP{0.9} & \relativeimpP{1.7} & \relativeimpP{0.9} & \relativeimpP{1.3} & \relativeimpP{0.5} & \relativeimpP{1.9} & \relativeimpP{3.3} & \relativeimpP{5.6} & \relativeimpN{0.4} & \relativeimpP{0.4} & \relativeimpP{1.0} & \relativeimpP{3.7} & \relativeimpP{3.1} & \relativeimpP{3.6} & \relativeimpP{3.4} & \relativeimpP{0.4} & \relativeimpP{18.5} & \relativeimpP{0.7} \\
\bottomrule
\end{tabular}
}
\vspace{-2mm}
\end{table*}

\renewcommand{\arraystretch}{0.9}
\begin{table*}[!t]
\centering
\caption{Semantic segmentation results on nuScenes~\cite{caesar2020nuscenes} in terms of mean Intersection over Union (mIoU), using OpenSeg feature extraction. 
\ourmethod outperforms the OpenScene on 11 out of 16 classes. Bold font indicates best performance.
Key: repr.: reproduced by us.
}
\label{tab:outdoor_sem_seg}
\vspace{-3mm}
\tabcolsep 5pt
\resizebox{\linewidth}{!}{%
\begin{tabular}{l|c| c c c c c c c c c c c c c c c c}
\toprule
& mIoU & \rotatebox{45}{Barr.} & \rotatebox{45}{Bicyc.} & \rotatebox{45}{Bus} & \rotatebox{45}{Car} & \rotatebox{45}{Const.} & \rotatebox{45}{Motor.} & \rotatebox{45}{Pers.} & \rotatebox{45}{Traf. c.} & \rotatebox{45}{Trail.} & \rotatebox{45}{Truck} & \rotatebox{45}{Dr. Sur.} & \rotatebox{45}{Oth. Fl.} & \rotatebox{45}{Sidew.} & \rotatebox{45}{Terr.} & \rotatebox{45}{Manm.} & \rotatebox{45}{Veget.}\\
\midrule
OpenScene (repr.) & 20.6 & 1.9 & \bf17.1 & 31.1 & 28.6 & 14.8 & \bf20.9 & \bf18.5 & 5.1 & 13.9 & 19.6 & 34.7 & 0.3 & \bf15.2 & 23.3 & \bf39.5 & 45.8 \\
\ourmethod (ours) & \bf22.6 & \bf2.0 & 16.3 & \bf39.2 & \bf35.8 & \bf15.2 & 18.7 & 12.9 & \bf7.1 & \bf15.7 & \bf25.9 & \bf49.9 & \bf1.3 & 14.9 & \bf27.0 & 33.2 & \bf47.3 \\
\midrule
$\Delta_\text{IoU}$ & \relativeimpP{2.0} & \relativeimpP{0.1} & \relativeimpN{0.8} & \relativeimpP{8.1} & \relativeimpP{7.2} & \relativeimpP{0.4} & \relativeimpN{2.2} & \relativeimpN{5.6} & \relativeimpP{2.0} & \relativeimpP{1.8} & \relativeimpP{6.3} & \relativeimpP{15.2} & \relativeimpP{1.0} & \relativeimpN{0.3} & \relativeimpP{3.7} & \relativeimpN{6.3} & \relativeimpP{1.5} \\
\bottomrule
\end{tabular}
}
\end{table*}

\subsection{Semantic Scene Segmentation}\label{sec:exp:sceneseg}
\noindent\textbf{Setting.}~We evaluate \ourmethod on ScanNet~\cite{dai2017scannet} and nuScenes~\cite{caesar2020nuscenes} for the semantic scene segmentation task.
ScanNet is an \mbox{RGBD} dataset including 2.5M views of 1,513 indoor scenes, annotated with point-level semantic labels and comprising 20 object classes.
We evaluate methods on the original validation set.
nuScenes is a LiDAR dataset of 400K outdoor scene point clouds with semantic annotations for 16 different classes. 
We apply \ourmethod to the 3D point-level features extracted with different 2D feature extractors, specifically those used in OpenScene~\cite{Peng2023}, based on OpenSeg and LSeg, and that used in ConceptFusion~\cite{Jatavallabhula2023}.

\noindent\textbf{Results.}~In Tab.~\ref{tab:indoor_sem_seg} we report the original results of OpenScene~\cite{Peng2023} and the ones reproduced by us with the \textit{``2D Fusion''}-version of their approach in the indoor dataset ScanNet.
\textit{2D Fusion} is the only training-free version of OpenScene.
For ConceptFusion~\cite{Jatavallabhula2023}, we reproduced the results with the original code\footnote{These results are different from those reported in the original paper \cite{Jatavallabhula2023} because authors only evaluated a subset of the classes and scenes.}.
\ourmethod outperforms OpenScene (OpenSeg), OpenScene (LSeg), and ConceptFusion by $+2.0$, $+3.9$, and $+3.1$ mIoU on average, respectively. 
Specifically, \ourmethod outperforms OpenScene (OpenSeg) on 17 out 20 classes of ScanNet, with significant gains on \textit{wall}~(+5.8),  \textit{floor}~(+8.7), and \textit{bathtub}~(+6.2) classes.
\ourmethod outperforms OpenScene (LSeg) on all the 20 classes, in particular for \textit{floor}~(+7.5), \textit{bathtub}~(+6.2), and  \textit{toilet}~(+9.6) classes.
\ourmethod outperforms ConceptFusion on 19 out of 20 classes, with a significant margin in the case of \textit{bathtub}~(+18.5), \textit{books}~(+5.6), and \textit{floor}~(+5.2) classes.

\begin{figure*}[!hbt]
\centering
\includegraphics[width=0.9\textwidth]{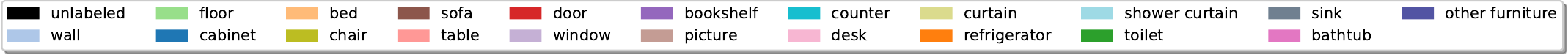}
\includegraphics[width=0.16\textwidth]{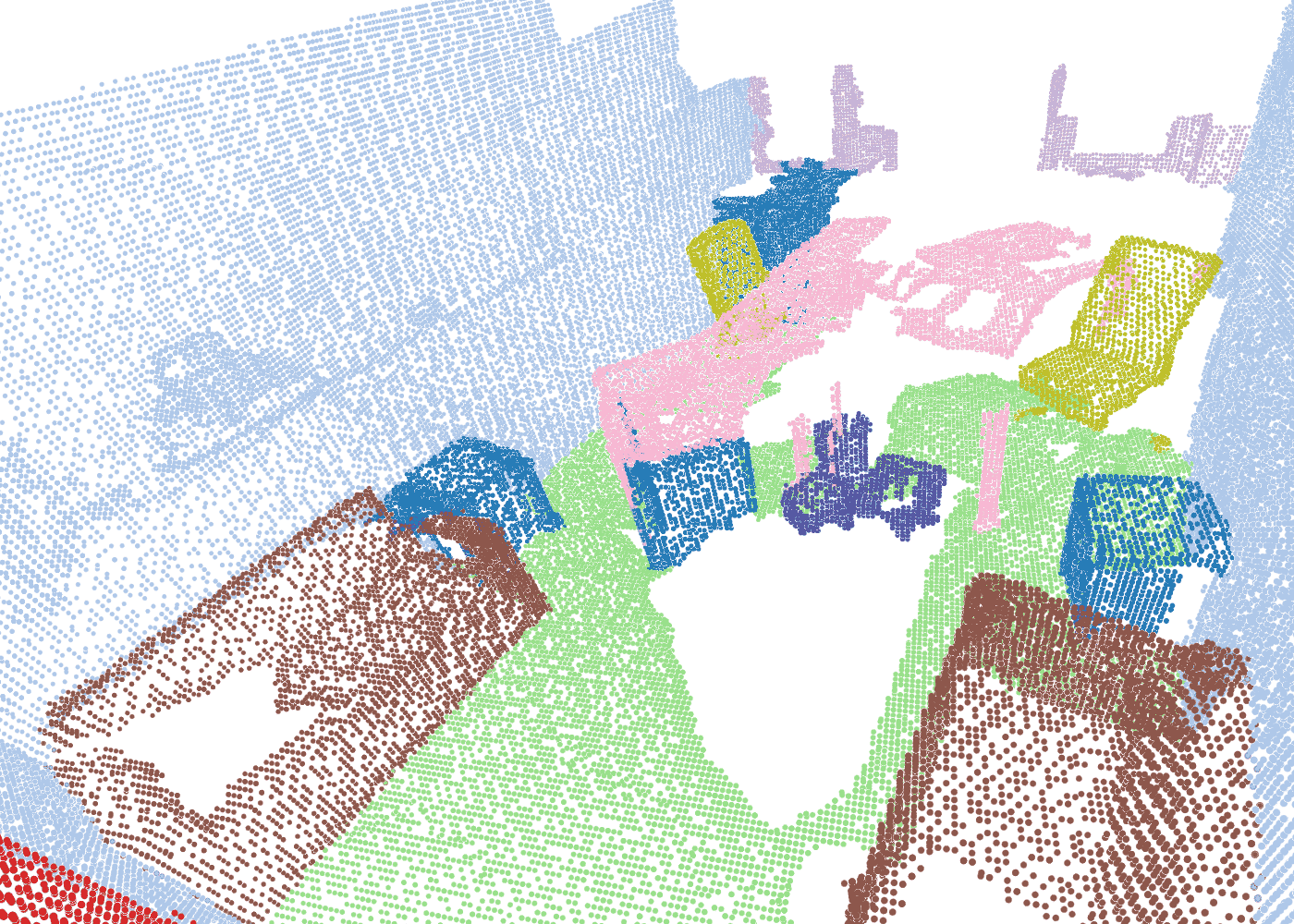}
\put(-86,15){\color{black}\footnotesize \rotatebox{90}{\textbf{GT}}}
\includegraphics[width=0.16\textwidth]{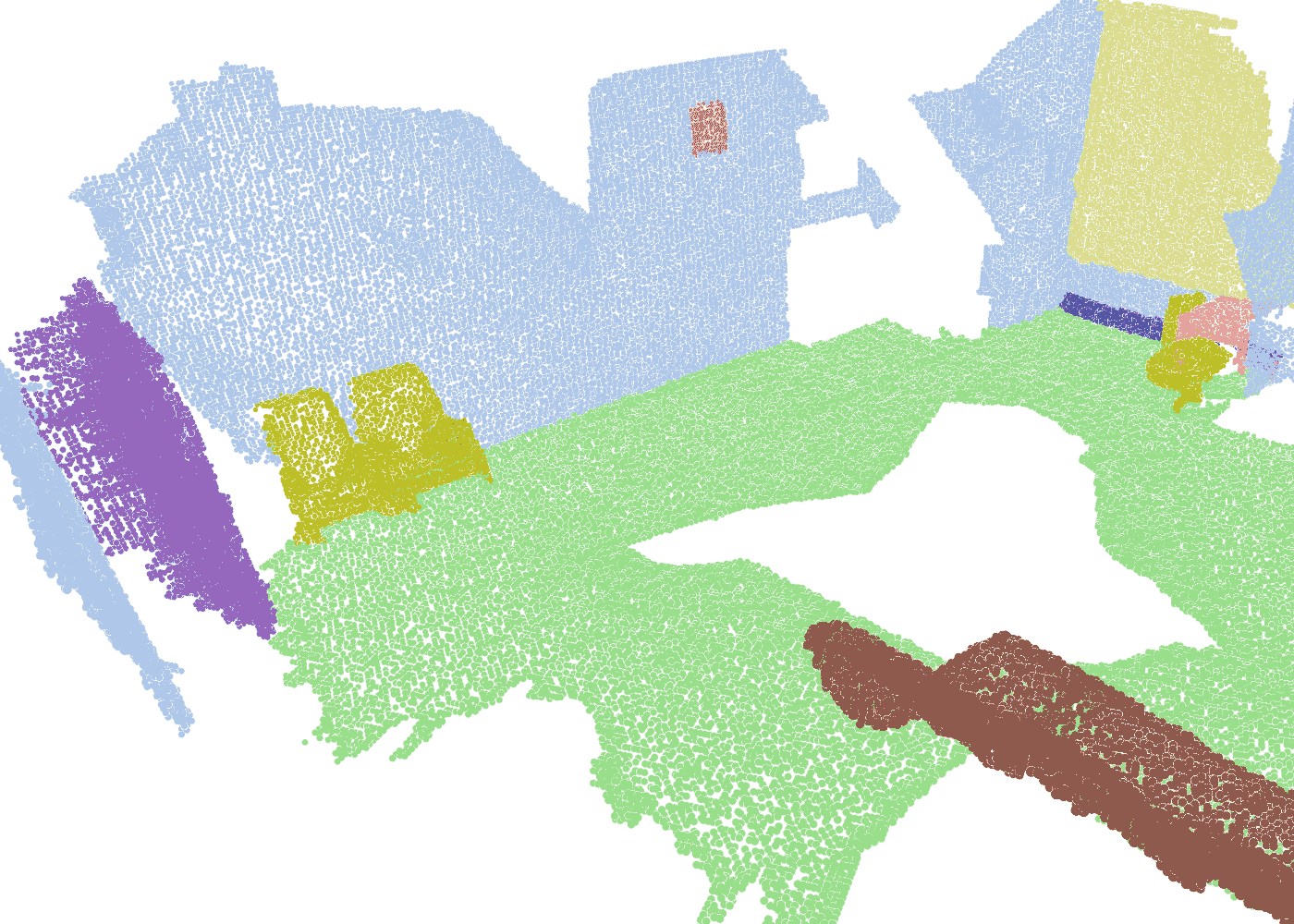}
\includegraphics[width=0.16\textwidth]{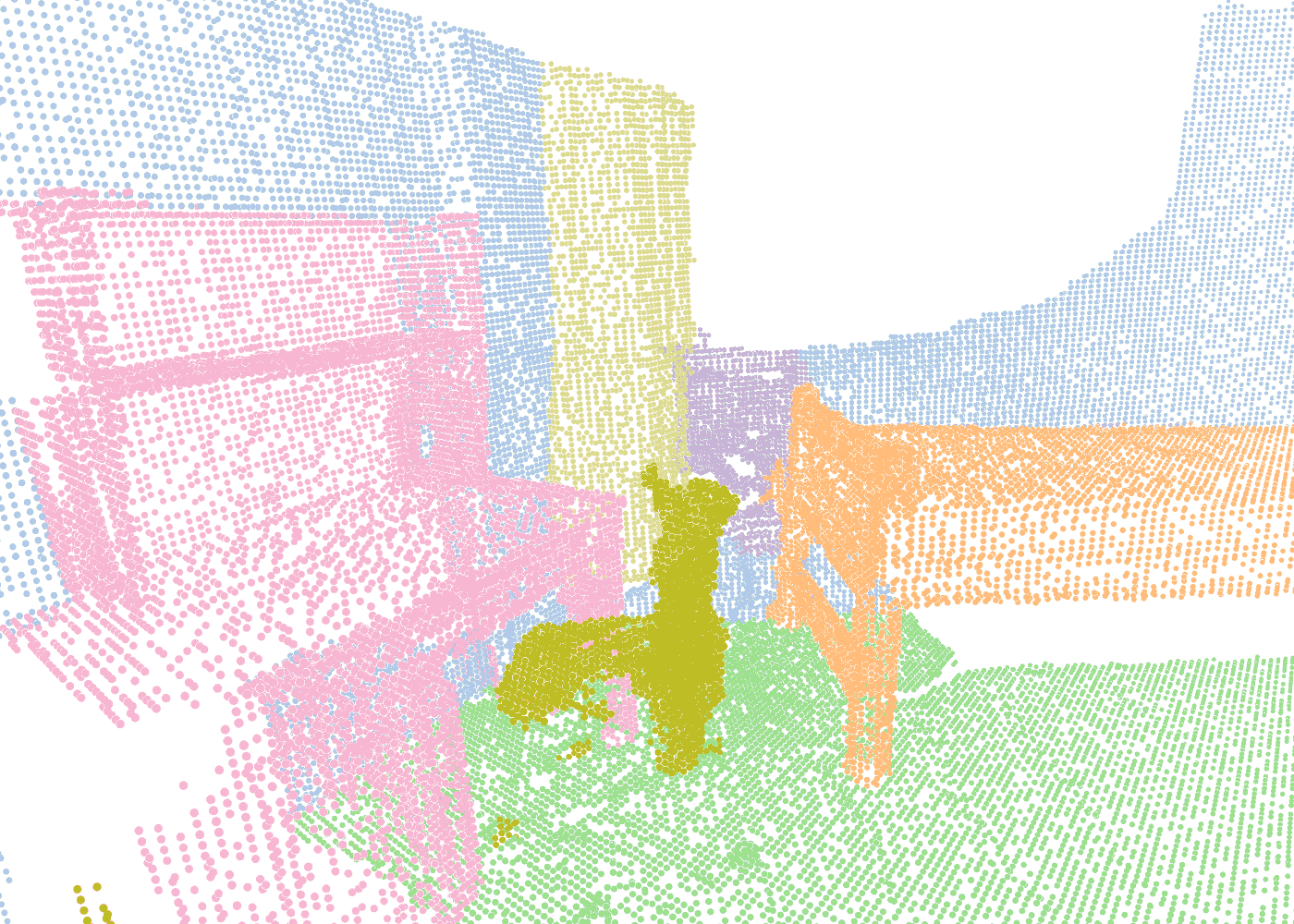}
\includegraphics[width=0.16\textwidth]{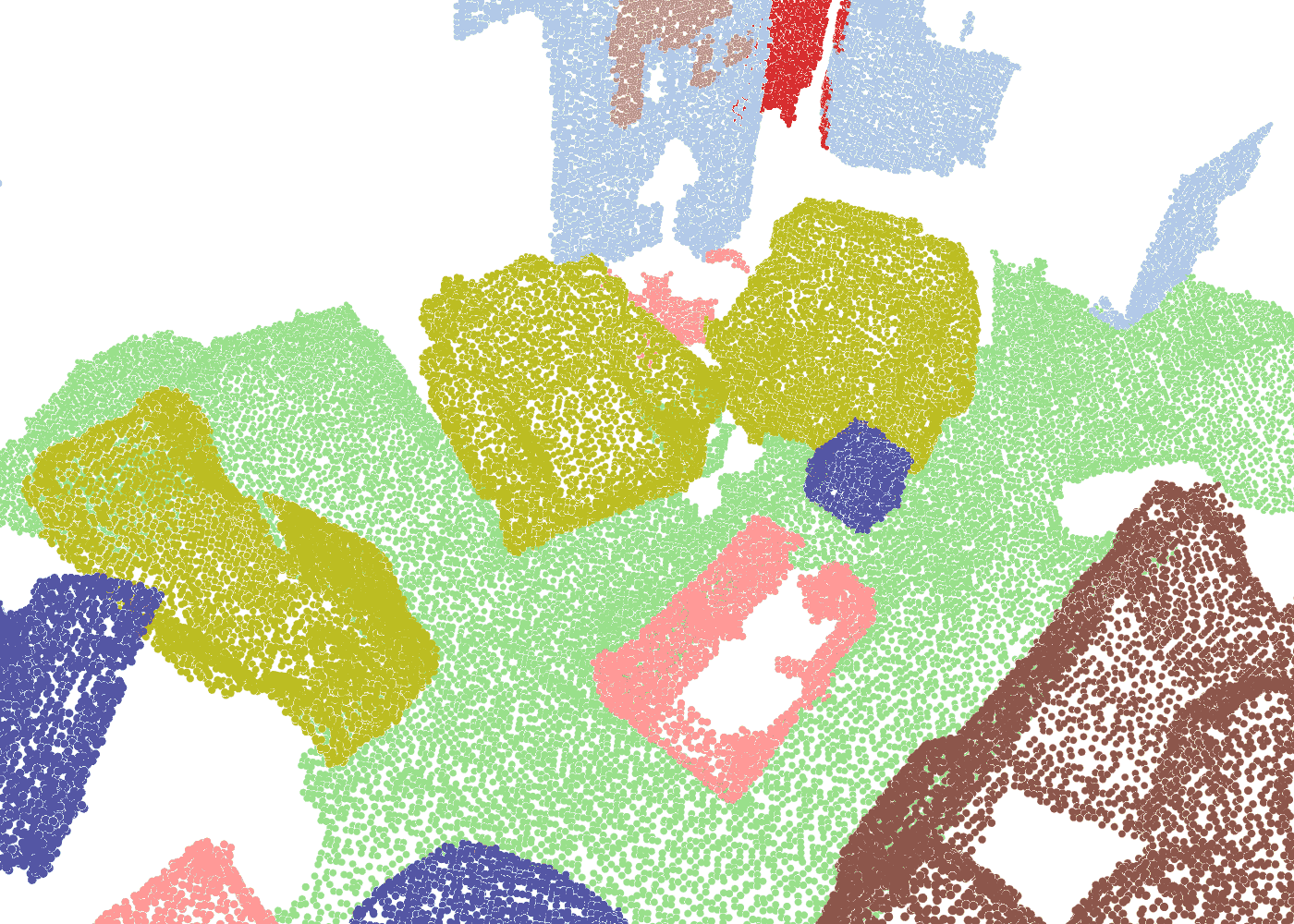}
\includegraphics[width=0.16\textwidth]{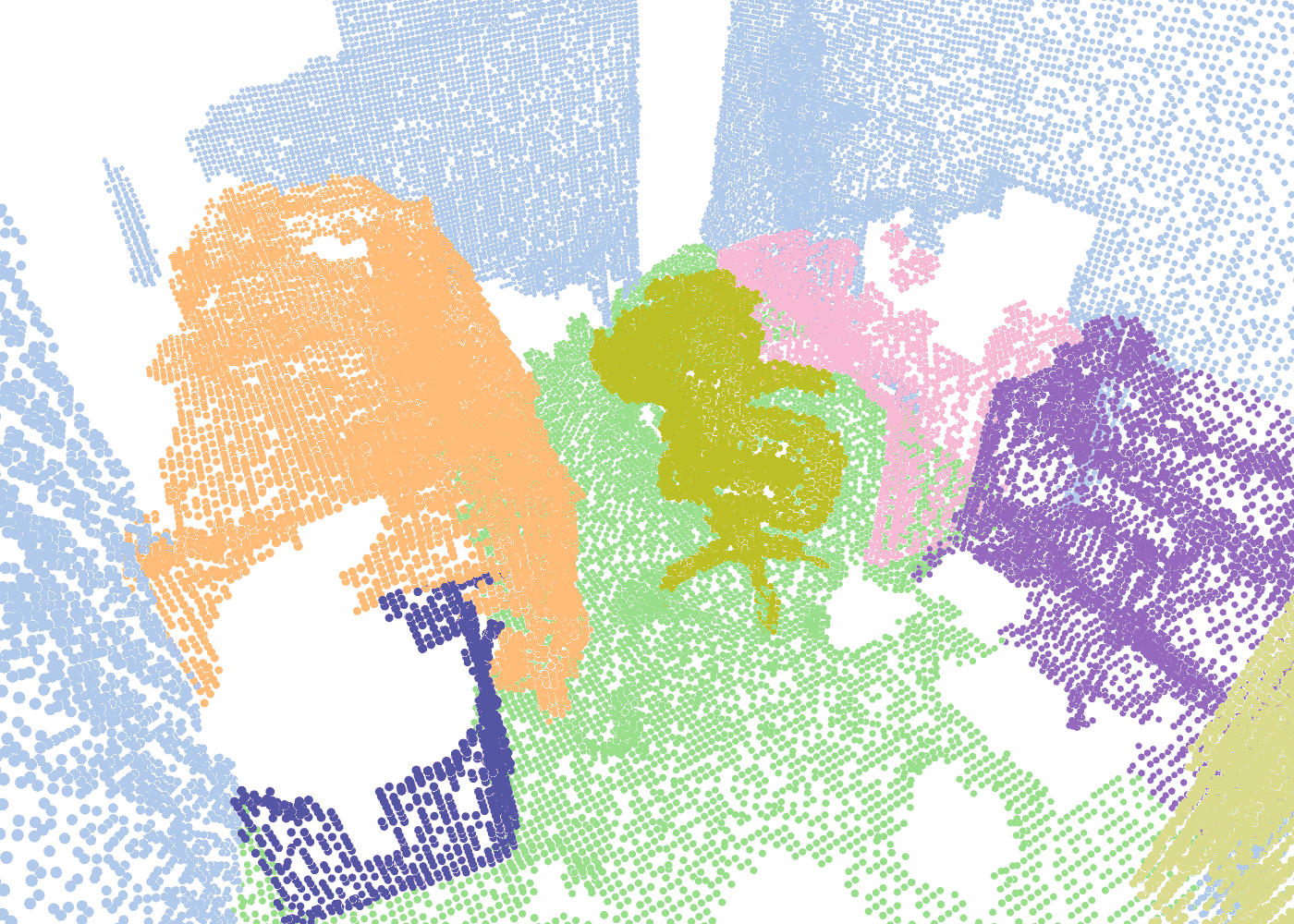}
\includegraphics[width=0.16\textwidth]{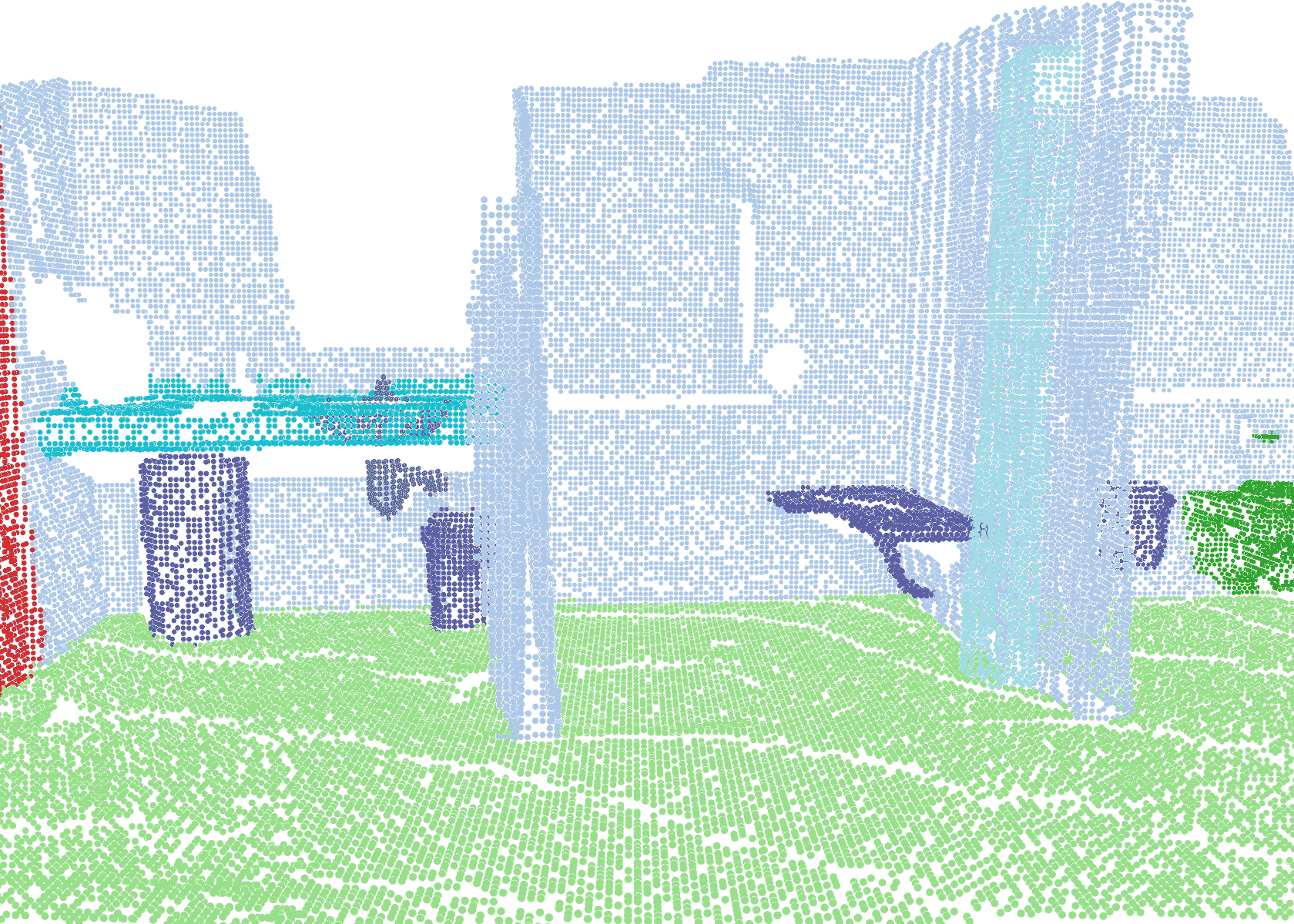}
\includegraphics[width=0.16\textwidth]{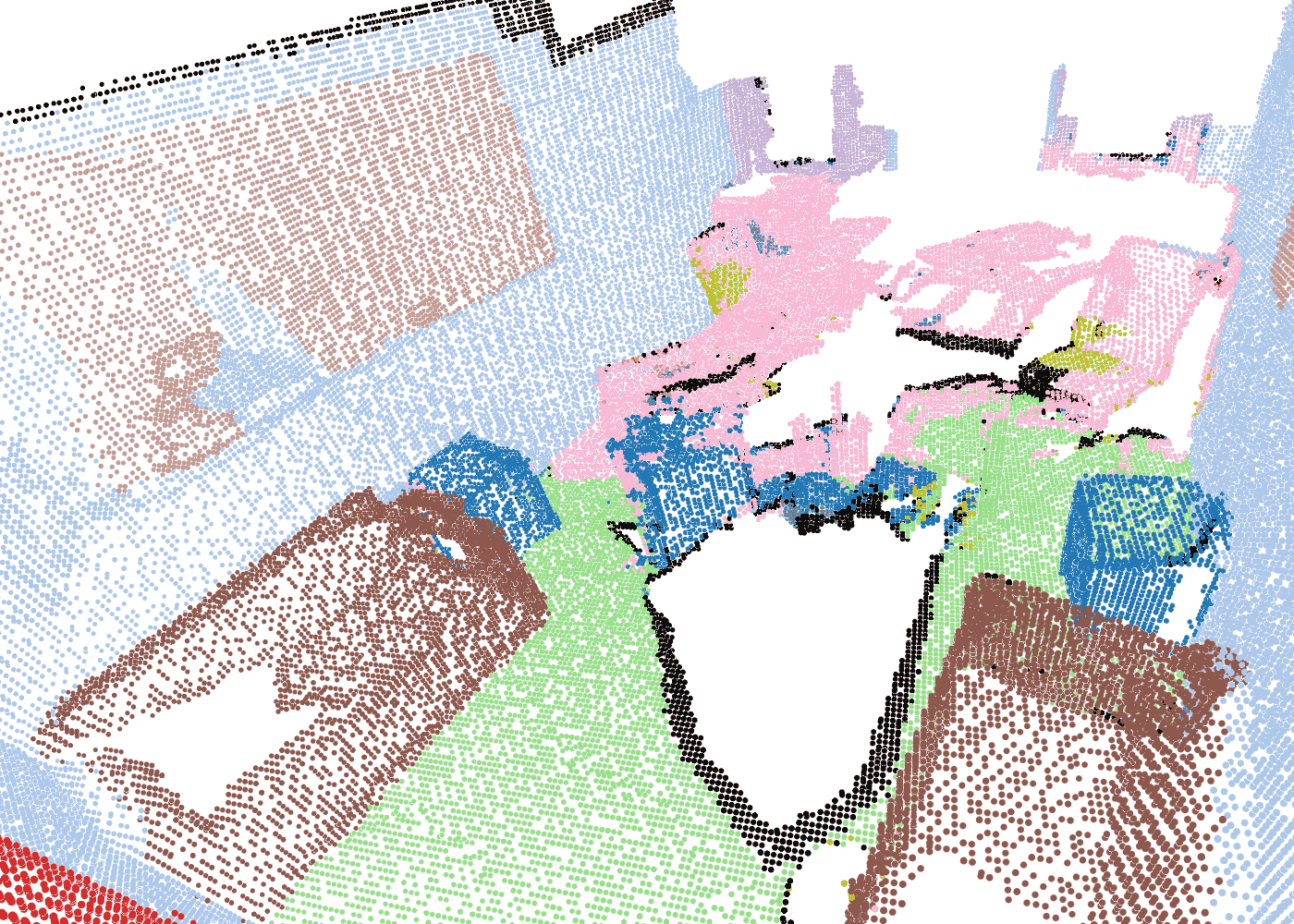}
\put(-86,12){\color{black}\footnotesize \rotatebox{90}{\textbf{OpenScene}}}
\includegraphics[width=0.16\textwidth]{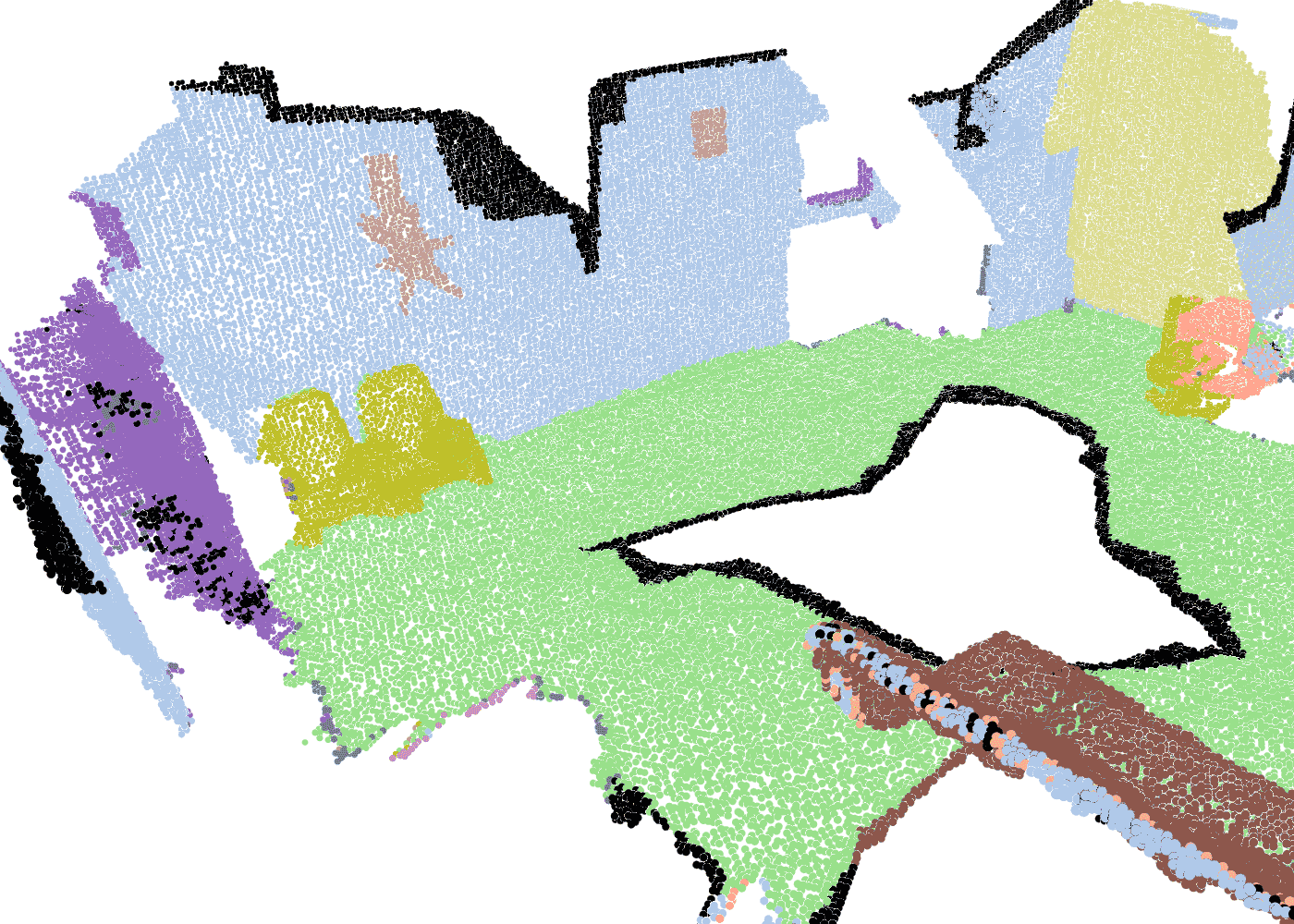}
\includegraphics[width=0.16\textwidth]{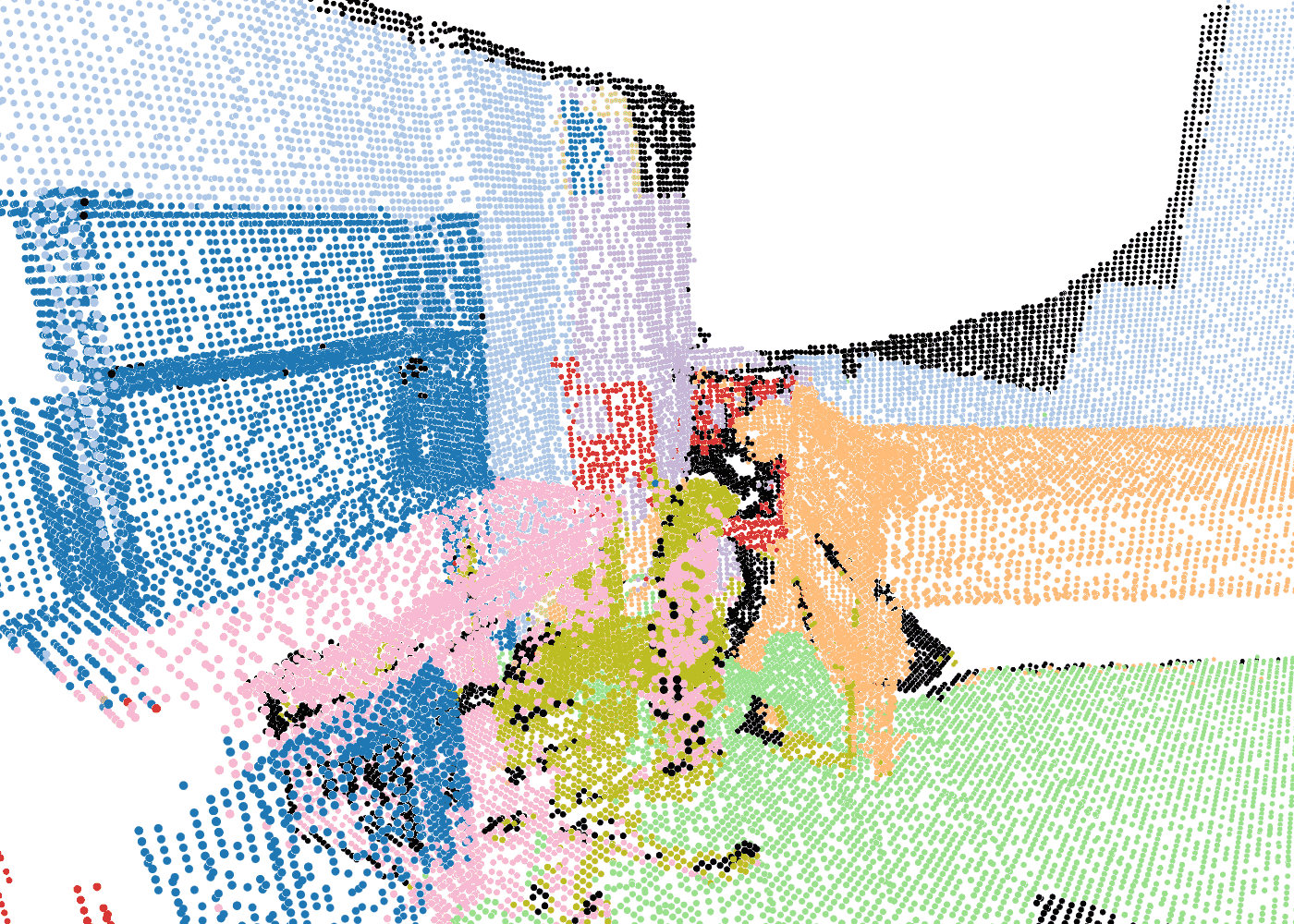}
\includegraphics[width=0.16\textwidth]{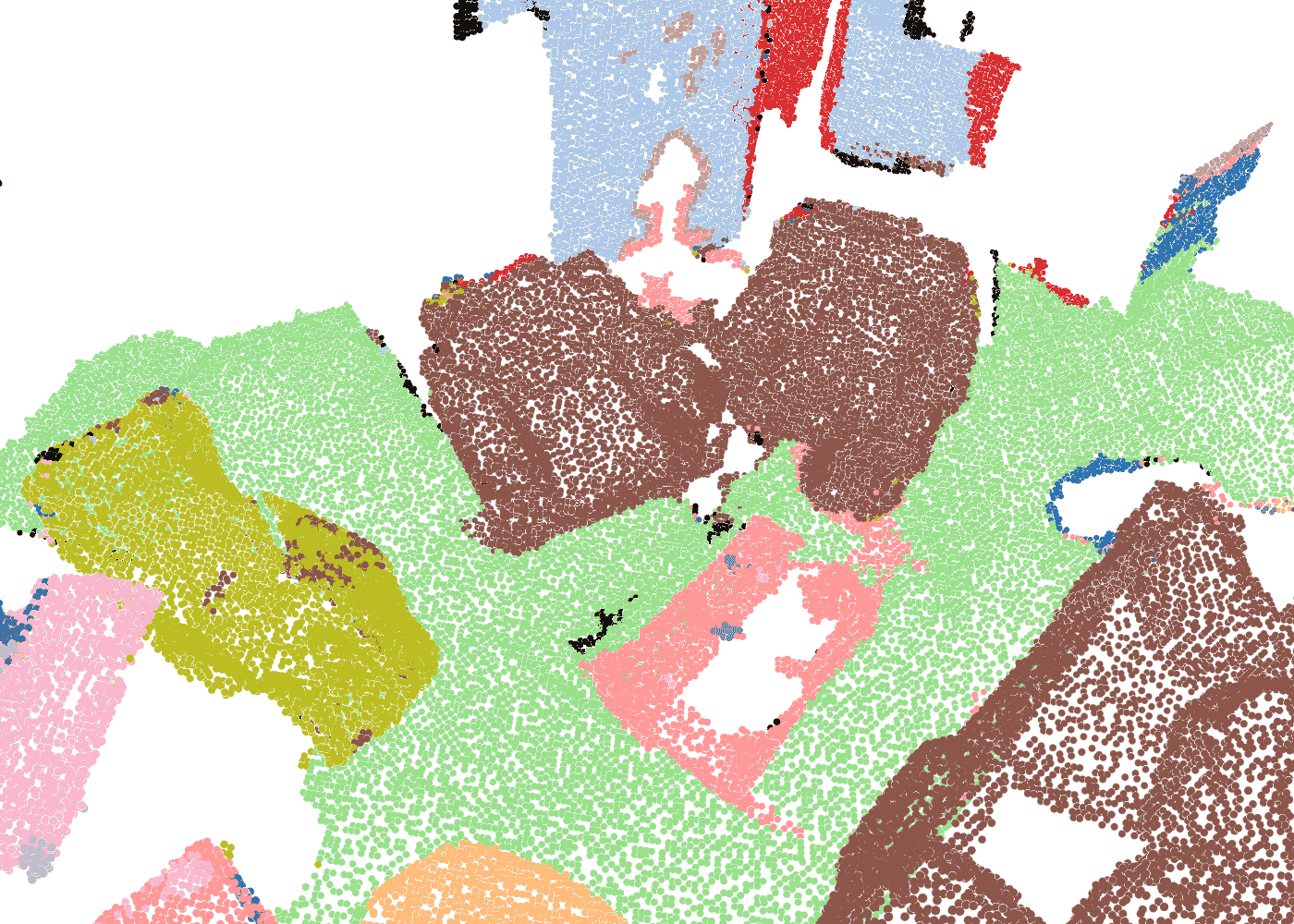}
\includegraphics[width=0.16\textwidth]{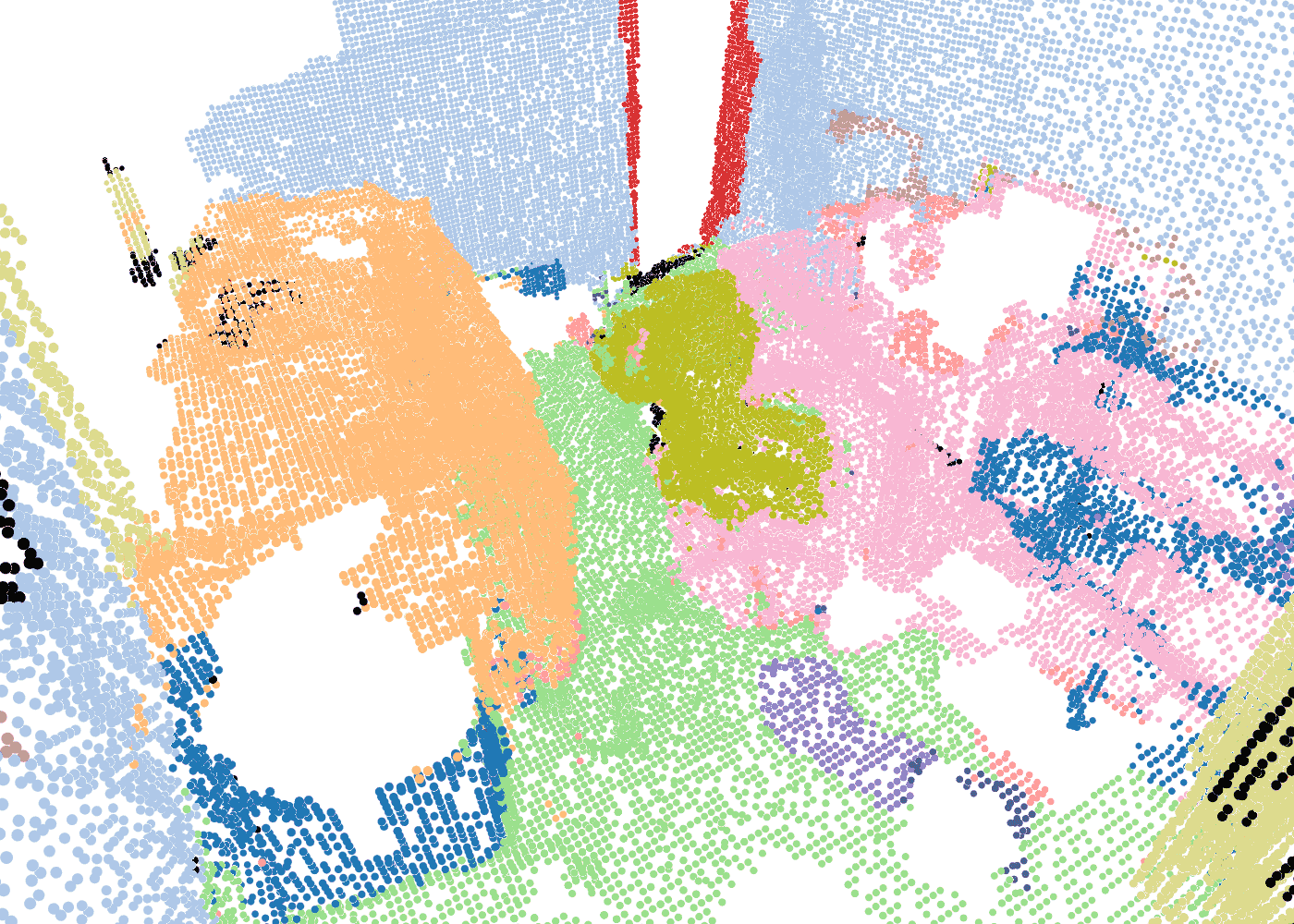}
\includegraphics[width=0.16\textwidth]{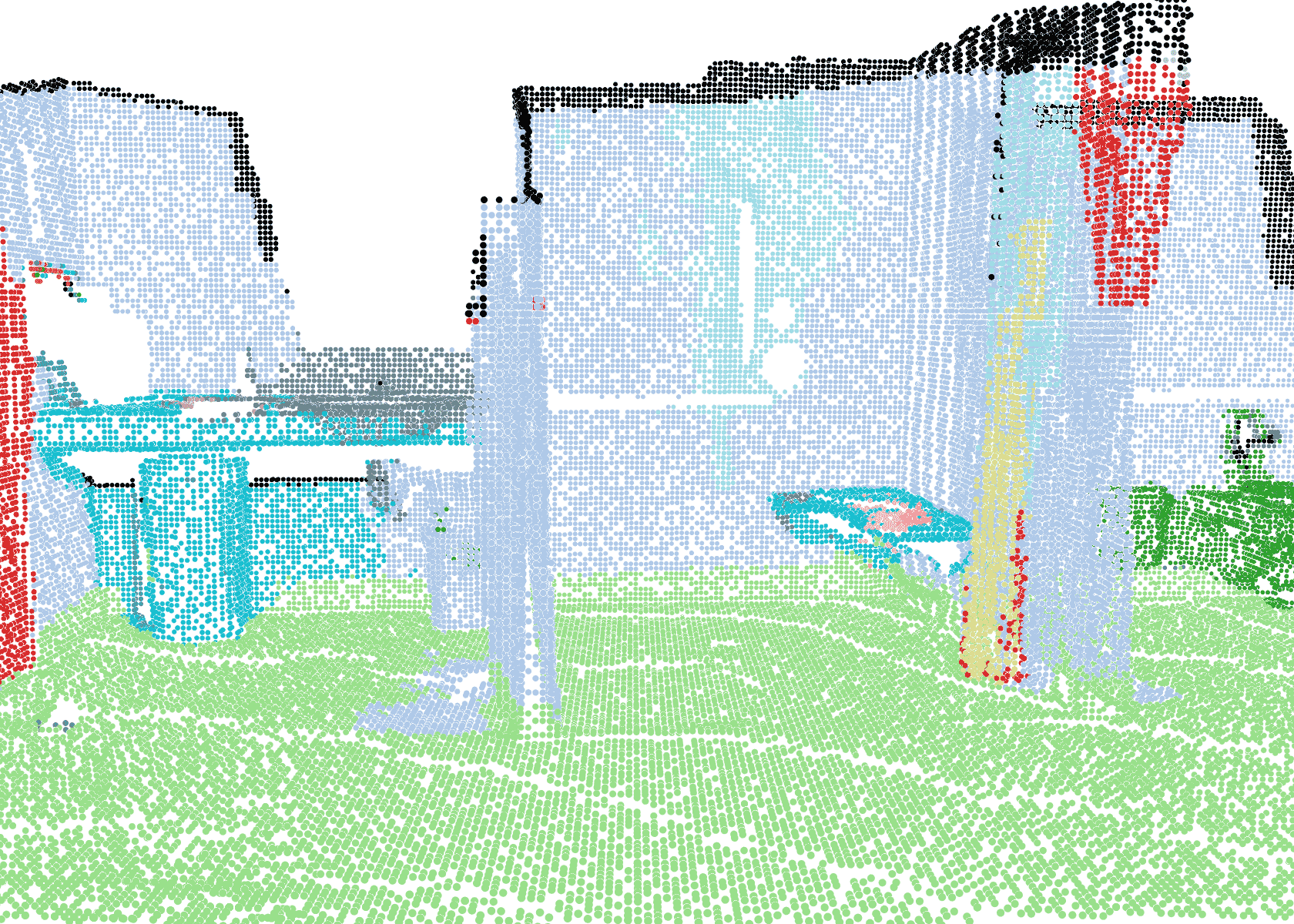}
\includegraphics[width=0.16\textwidth]{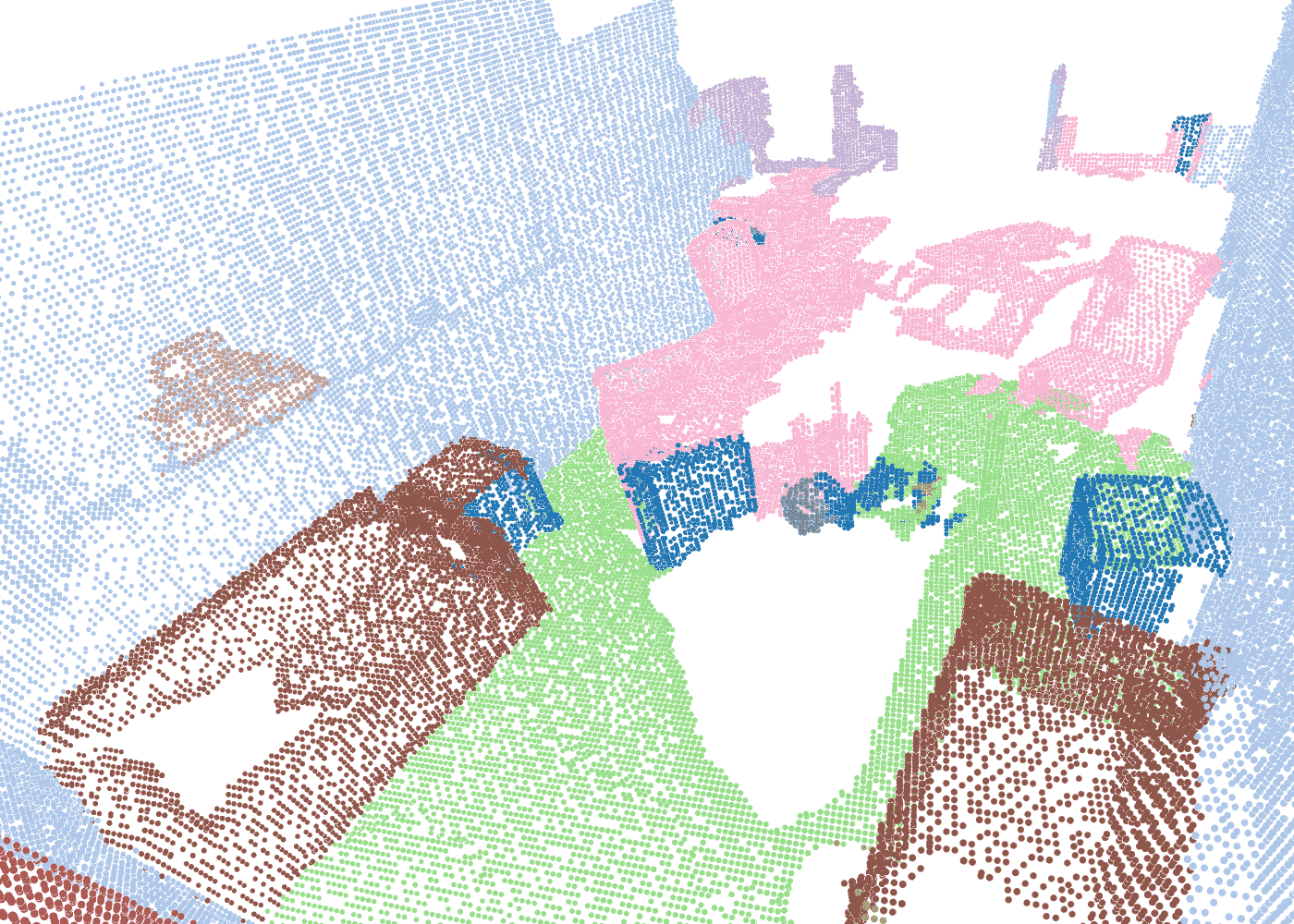}
\put(-86,15){\color{black}\footnotesize \rotatebox{90}{\textbf{\ourmethod}}}
\includegraphics[width=0.16\textwidth]{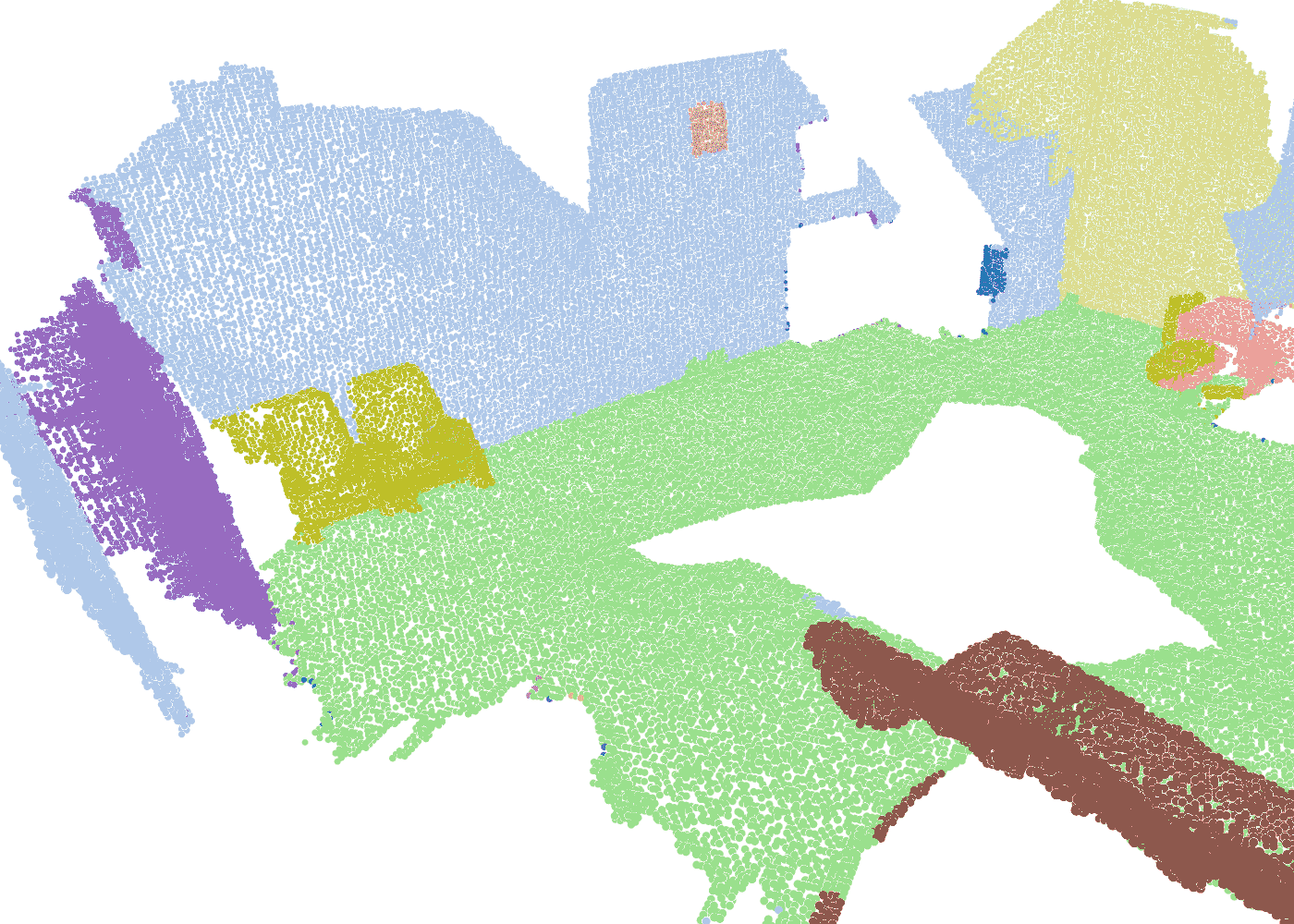}
\includegraphics[width=0.16\textwidth]{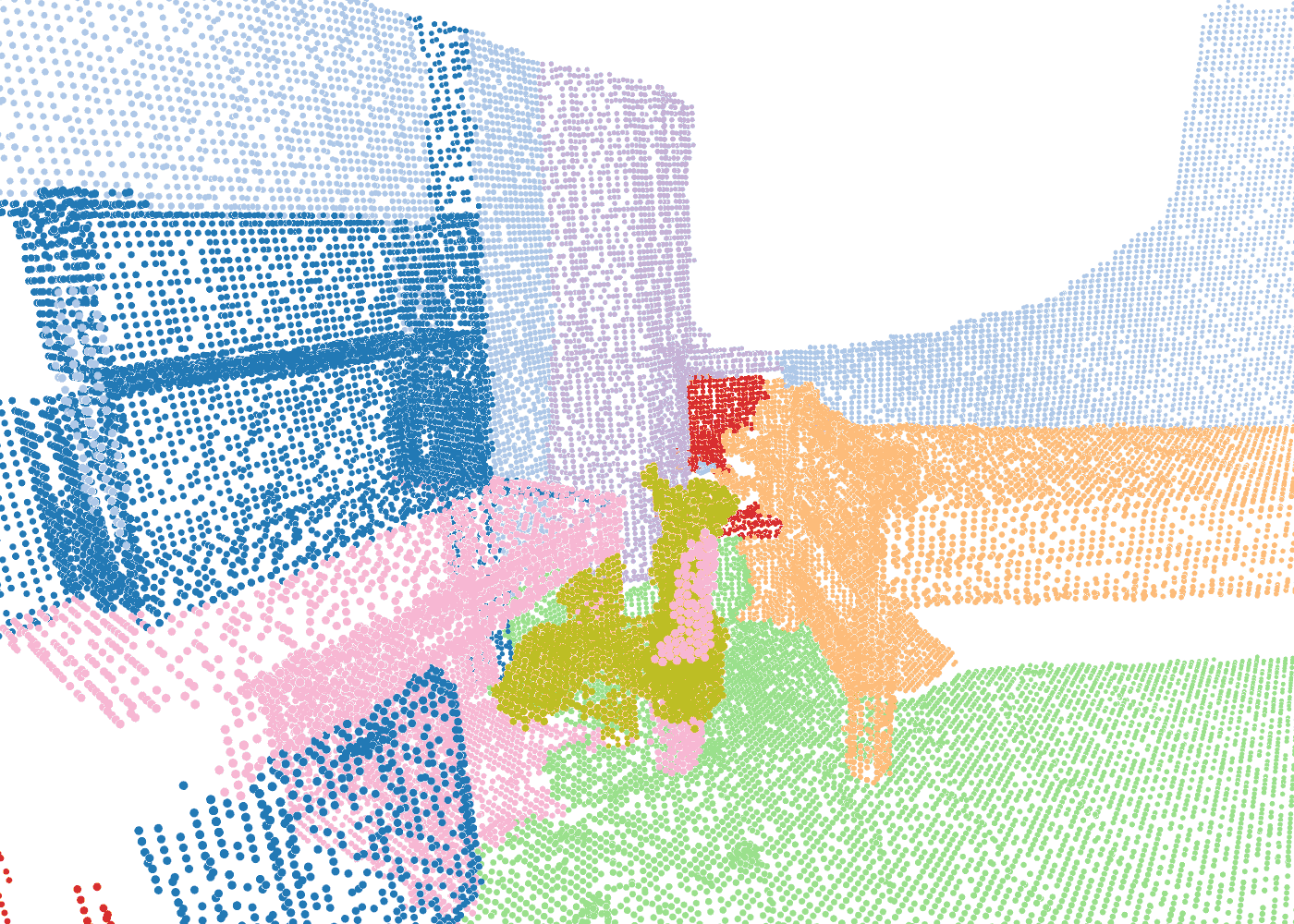}
\includegraphics[width=0.16\textwidth]{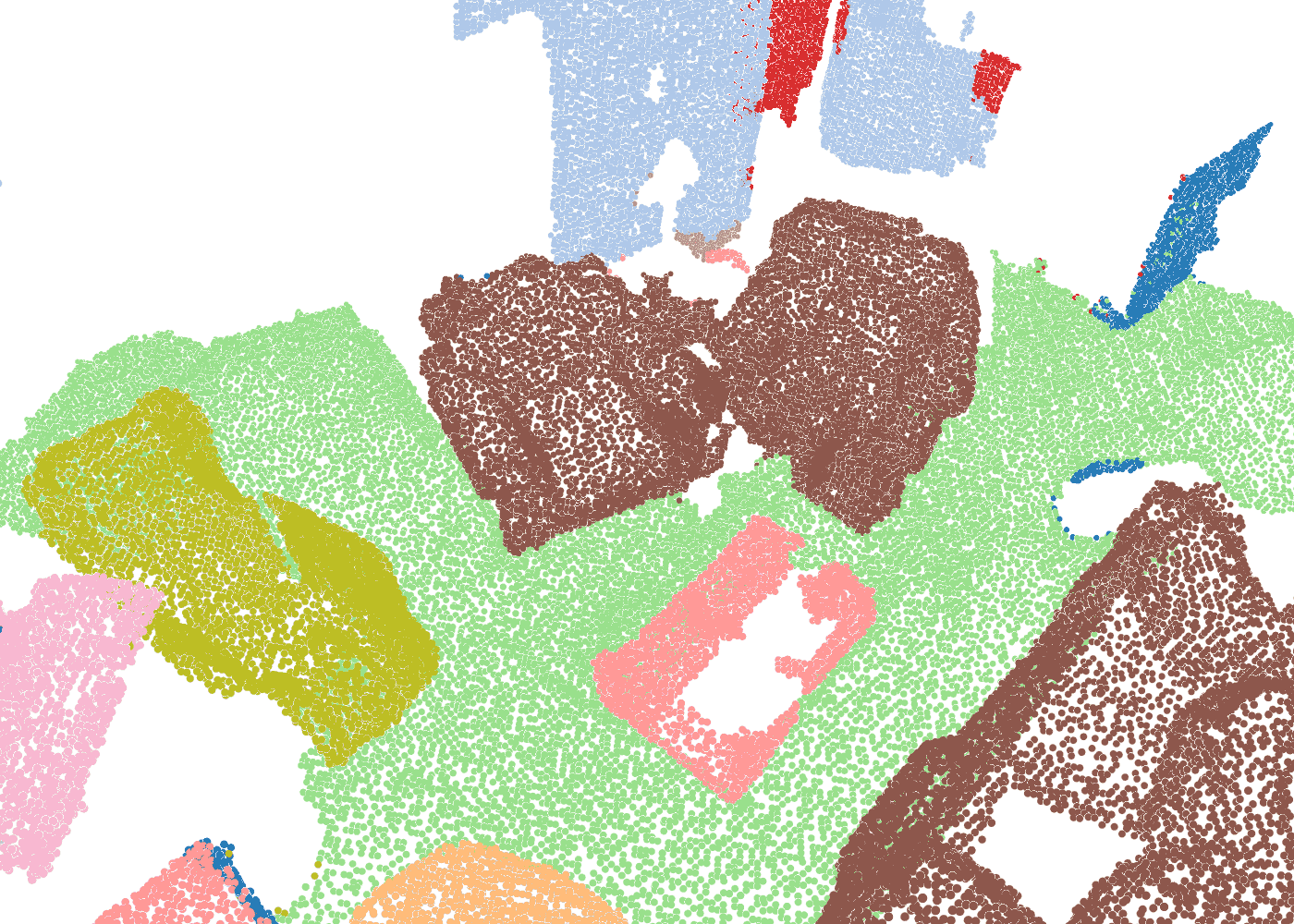}
\includegraphics[width=0.16\textwidth]{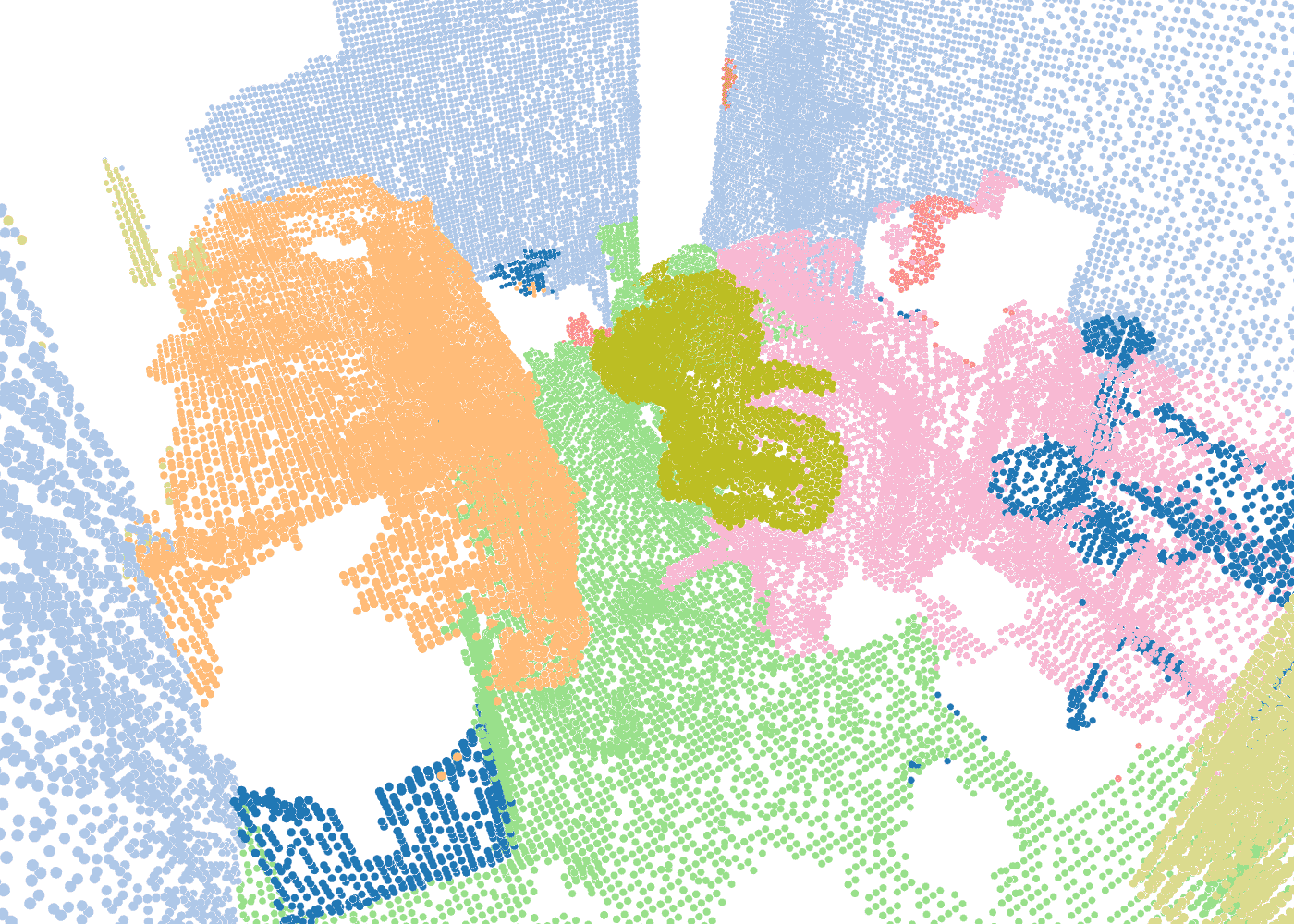}
\includegraphics[width=0.16\textwidth]{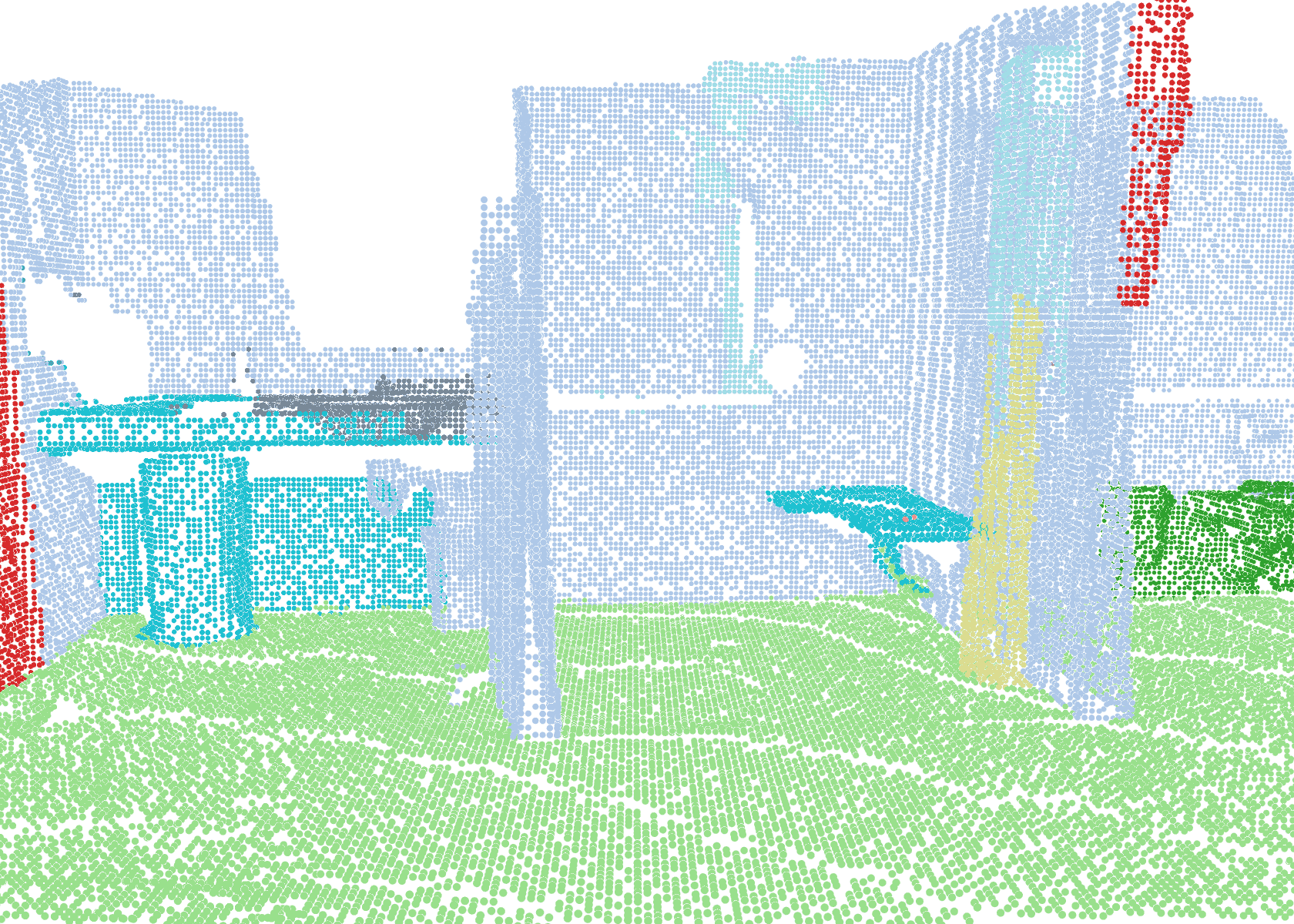}
\vspace{-3mm}
\caption{Zero-shot semantic segmentation results on ScanNet~\cite{dai2017scannet} using OpenSeg feature extraction (Tab.~\ref{tab:indoor_sem_seg}).
(top row) ground-truth annotations, (middle row) OpenScene (OpenSeg), and (bottom row) \ourmethod.}
    \label{fig:sem_seg}
    \vspace{-4mm}
\end{figure*}

Tab.~\ref{tab:outdoor_sem_seg} reports the results in the outdoor dataset nuScenes. 
Similarly to Tab.~\ref{tab:indoor_sem_seg}, we reproduce the results of the \textit{2D Fusion}-version of OpenScene~\cite{Peng2023} using the OpenSeg feature extractor as it is the only one available.
\ourmethod outperforms OpenScene (OpenSeg) by 2.0 mIoU on average, and on 11 out of the 16 classes, with a improvement on \textit{bus} by 8.1 mIoU, \textit{truck} by 6.3 mIoU, and \textit{drivable surface} by 15.2 mIoU.
Fig.~\ref{fig:sem_seg} reports examples of qualitative results.
\ourmethod produces more homogeneous segmented regions and reduces noise, particularly at the point cloud boundaries.


\subsection{Ablation study}\label{sec:exp:ablation}

We use the part segmentation task on ShapeNetPart for the ablation study of \ourmethod.
The Supplementary Material contains additional ablation studies.

\begin{figure}
    \centering
    \includegraphics[width=1\linewidth]{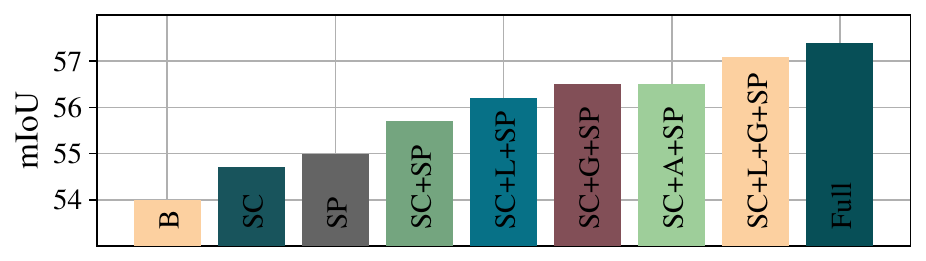}
    \vspace{-8mm}
    \caption{Ablation study on \ourmethod's modules, tested on the part segmentation task.
Keys: B: baseline. SC: superpoint computation. SP: superpoint-to-point aggregation. G: global aggregation. L: local aggregation. A: anchor projection. 
Full: the whole model.}
\label{tab:ab_lg}
\vspace{-3mm}
\end{figure}

\noindent\textbf{Effectiveness of aggregation levels.}~We assess the contributions of the local, global, and superpoint-to-point aggregations, and superpoint computation and anchor projection steps: for convenience, we name these steps as L, G, SP, SC and A.
Since \ourmethod first clusters a point cloud into superpoints and then propagates them back to its original resolution, we adopt the propagation strategy outlined in \cite{qi2017pointnet++} as our baseline, in short B. 
Fig.~\ref{tab:ab_lg} reports the results of our ablation study. 
B scores 54.0 mIoU.
Superpoint computation (SC) and superpoint-to-point aggregation (SP) improve the baseline by $0.7$ and $1.0$ mIoU, respectively.
The combination SC+SP results in a gain of $1.7$ mIoU.
Local aggregation (SC+L+SP) leads to an improvement of $0.5$ mIoU over SC+SP, underlining the importance of local smoothing.
Global aggregation (SC+G+SP) improves $0.8$ mIoU over SC+SP, indicating the positive effect of global semantic context.
Anchor projection (SC+A+SP) improves $0.8$ mIoU over SC+SP, while the combination of SC+L+G+SP results in $57.1$ mIoU. 
The full \ourmethod reaches $57.4$ mIoU.

\noindent\textbf{Impact of feature types.}~Tab.~\ref{tab:ab_geo} reports the results of the different types of features in the aggregation process. 
Our experiments indicate that incorporating geometric, semantic and coordinate information significantly enhances the effectiveness of local aggregation. 
Similarly, global aggregation based on geometric and semantic features allows for a great increase in performance.
It improves the segmentation accuracy by nearly +1.2 mIoU (55.7\% vs. 56.9\%).

\renewcommand{\arraystretch}{0.9}
\begin{table}[t]
\small
\centering
\caption{Ablation study evaluating the influence of geometric information on part segmentation within the ShapeNetPart dataset. 
Keys: Coors: coordinate, CLIP: \vlmfeat, GA: global aggregation, LA: local aggregation.
}
\label{tab:ab_geo}
\vspace{-2mm}
\tabcolsep 6pt
\resizebox{\columnwidth}{!}
{%
\begin{tabular}{l|l| c c c c c c c c}
\toprule
& Feature & \multicolumn{7}{c}{Feature Combinations} \\
\midrule
\multirow{3}{*}{LA} 
& CLIP \cite{zhu2023pointclip} & \checkmark & \checkmark & \checkmark & \checkmark & & &  &\checkmark\\
& Coors & ~ & \checkmark  &  & \checkmark & & & &\checkmark \\
& FPFH \cite{rusu2009fast} & ~ & ~ & \checkmark & \checkmark & & & &\checkmark \\
\midrule
\multirow{2}{*}{GA} 
& CLIP \cite{zhu2023pointclip}  & ~ & ~ & & & \checkmark &  &\checkmark  & \checkmark\\
& FPFH \cite{rusu2009fast} & ~ & ~ & ~ &  & & \checkmark & \checkmark & \checkmark\\
\midrule
& mIoU & 55.7 & 56.2 & 56.2 & 56.4 & 55.7 & 56.5 & 56.9  & 57.4 \\
\bottomrule
\end{tabular}
}
\vspace{-3mm}
\end{table}

\subsection{Computational analysis}
\vspace{-2mm}
We compare the processing time of \ourmethod against the aggregation approaches of the evaluated methods and report the average time increment per point cloud.
We run the baselines methods and \ourmethod~on a NVIDIA A40 48GB GPU and report inference time in the Tab.~\ref{tab:time}.
Although we did not implement any low-level optimization of our algorithm, we can observe that the increment introduced by \ourmethod on top of PointCLIPv2 and OpenScene is limited.
\begin{table}[h!]
\centering
\tabcolsep 10pt
\captionsetup{skip=4pt} 
\caption{Comparison of inference times across different methods.}
\label{tab:time}
\vspace{-1mm}
\resizebox{\columnwidth}{!}
{%
\begin{tabular}{c|cc|cc}
\toprule
~  & \multicolumn{2}{c|}{ShapeNet} & \multicolumn{2}{c}{ScanNet} \\
\cline{2-5}
& PointCLIPv2~\cite{zhu2023pointclip} & \ourmethod & OpenScene~\cite{Peng2023} & \ourmethod \\
\midrule
Time(ms)  & 7.52  & 9.83 & 2088.72 & 2125.61 \\
\bottomrule
\end{tabular}
}
\vspace{-4mm}
\end{table}

\section{Conclusions}
\vspace{-2mm}
We presented the first training-free aggregation technique that leverages
the point cloud’s 3D geometric structure to improve the quality of the transferred \vlmfeats.
\ourmethod is designed to iteratively aggregate representations at local level first and global level then.
\vlmfeats are extracted from a CLIP model \cite{radford2021learning}, while \geofeats are extracted from the FPFH algorithm \cite{rusu2009fast}.
We carried out an extensive evaluation on three downstream tasks: classification, part segmentation, and semantic segmentation. 
We reported the results on both synthetic/real-world, and indoor/outdoor datasets, and showed that \ourmethod achieves new state-of-the-art results in all benchmarks.

\vspace{1mm}
\noindent\textbf{Limitations.}~\ourmethod depends on the quality of \vlmfeats, it may not accurately handle features of objects that constitute a small portion of the 3D scene, and it increases the inference time compared to a naive aggregation.

\noindent\textbf{Acknowledgement.} 
This work was supported by the PNRR project FAIR- Future AI Research (PE00000013), under the NRRP MUR program funded by the NextGenerationEU.

\clearpage

{
    \small
    \bibliographystyle{ieeenat_fullname}
    \bibliography{main}
}

\appendix
\clearpage

\twocolumn[
\centering
\begin{@twocolumnfalse}
\section*{Geometrically-driven Aggregation for Zero-shot 3D Point Cloud Understanding \\
Supplementary Material}
  \vspace{1cm} 
\end{@twocolumnfalse}
]

In this supplementary material, we provide implementation details in Sec.~\ref{sup:setting} and further experimental analyses in Sec.~\ref{sup:ad_result}.

\section{Implementation Details}\label{sup:setting}

\subsection{\vlmfeat extraction}
For classification tasks on both the ModelNet40~\cite{sharma2016vconv} and ObjectScanNN~\cite{uy2019revisiting} datasets, we adhere to the method used in PointCLIPv2~\cite{zhu2023pointclip}, sampling 1,024 points from each point cloud. These points are then projected into depth images from 10 different views for the extraction of \vlmfeats. 
As in PointCLIPv2~\cite{zhu2023pointclip}, we use the ViT-B/16~\cite{vaswani2017attention} model as the visual encoder within the CLIP framework, which comprises 12 layers of multi-head self-attention (MHSA). 
We extract the \vlmfeats corresponding to image patches during the attention process at the final MHSA layer. These \vlmfeats are then subjected to bilinear interpolation, a technique we utilize to upscale the \vlmfeats to the original size of the image, specifically to $224 \times 224$. 
Given that a single view projection captures only a partial point cloud, we utilize multi-view back projection to ensure thorough predictions for all points in the cloud. For points observable from multiple views, we perform linear interpolation of the \vlmfeats, fine-tuning this process based on the varying weights of the different views. These \vlmfeats are then fed into our \ourmethod to obtain enhanced point-level representations. Subsequently, max-pooling is applied to generate the global features. Additionally, we extract global features using the ViT-B/16 model from each view, consistent with PointCLIPv2.

For the part segmentation task on the ShapeNetPart~\cite{yi2016scalable} dataset, we sample 2,048 points per point cloud. These points are then projected onto depth images from 10 different viewpoints for \vlmfeat extraction using the ViT-B/16 model. After extraction, these \vlmfeats undergo bilinear interpolation for upsampling to their original image size. Similar to the classification process, we apply multi-view back projection to the point cloud using the \vlmfeats. These \vlmfeats are subsequently processed through our \ourmethod to achieve enhanced point-level representations.

In assessing the semantic segmentation performance on the ScanNet~\cite{dai2017scannet} and nuScenes~\cite{caesar2020nuscenes} datasets, we utilize the \vlmfeats that are provided by OpenScene~\cite{Peng2023}. To ensure consistency and facilitate a fair comparison, we adhere to the standard voxel size of 0.02m as in OpenScene.
These \vlmfeats are then processed using \ourmethod, which facilitates the enhancement of point-level representations.

\subsection{Parameters}
Tab.~\ref{tab:parameter} presents the dataset-specific hyperparameters including the number of iterations (\(\Gamma\)), the number of superpoints (\(\bar{N}\)), the number of points (\(K_1\)) used for computing similarity, and the number of neighboring points (\(K_2\)) for local aggregation. 
Specifically, increasing the number of superpoints from \(\frac{N}{8}\) to a maximum of \(\bar{N} \leq \frac{N}{4}\) can slightly improve results, as denoted by \(\frac{N}{8} \leq \bar{N} \leq \frac{N}{4}\). $N$ is the number points of a point cloud.
Our experiments suggest that optimal accuracy is achieved when the number of superpoints is maintained between one-eighth and one-third of the original point cloud size. Balancing both accuracy and time complexity, the parameters we adopted in our experiments are as listed in Tab.~\ref{tab:parameter}.

\begin{table}[t]
    \centering
    \caption{Hyper-parameters configurations for different datasets.}
    \label{tab:parameter}
    \resizebox{1\columnwidth}{!}{
    \begin{tabular}{c|ccccc}
    \toprule
        Dataset  & Network & $\Gamma$ & $\bar{N}$ & $K_1 $ & $K_2$\\
        \midrule
        ModelNet40 & PointCLIPv2\cite{zhu2023pointclip} & 16 & 256 & 32 & 24 \\
        ObjectScanNN & PointCLIPv2\cite{zhu2023pointclip} & 16 & 256 & 32 & 24 \\
        ShapeNetPart & PointCLIPv2\cite{zhu2023pointclip} & 16 & 256 & 32 & 24 \\
        \midrule
        \multirow{3}{*}{ScanNet} & OpenSeg~\cite{ghiasi2022scaling} & 8 & 3000 & 48 & 32\\
                                 & LSeg~\cite{li2022languagedriven} & 8 & 3000 & 48 & 32\\
                                 & ConceptFusion\cite{Jatavallabhula2023} & 8 & 3000 & 48 & 32\\
        \midrule
        nuScenes & LSeg~\cite{li2022languagedriven} & 8 & 2400 & 48 & 32\\
        \bottomrule
    \end{tabular}
}
\end{table}

\subsection{Geometric representation (FPFH) extraction}
In our experiments, we compute FPFH~\cite{rusu2009fast} features for all points. To improve the computation time, we first downsample $M$ reference points. Then, we sample $K_3$ neighboring points for each point from the $M$ reference points within a radius $r_1$ for estimating the normals and $K_4$ neighboring points for each point from $M$ reference points within a radius $r_2$ for estimating FHFH. The details are reported in Tab.~\ref{tab:fpfh}.



\begin{table}[h]
    \centering
    \caption{Hyper-parameters configurations for FPFH computation on different datasets.}
    \label{tab:fpfh}
    \vspace{-2mm}
    \tabcolsep 4mm
    \resizebox{1\columnwidth}{!}{%
    \begin{tabular}{c|ccccc}
    \toprule
        Dataset & $\bar{M}$ & $K_3 $ & $K_4$  & $r_1$ & $r_2$\\
        \midrule
        ModelNet40 & 512 & 32 & 100 & 0.04 & 0.08 \\
        ObjectScanNN & 512 & 32 & 100 & 0.04 & 0.08 \\
        ShapeNetPart & 512 & 32 & 100 & 0.04 & 0.08 \\
        \midrule
        ScanNet & 4800 & 32 & 100 & 0.05 & 0.10\\
        \midrule
        nuScenes & 5200 & 32 & 100 & 0.05 & 0.10\\
        \bottomrule
    \end{tabular}
}
\end{table}

\section{Additional Experimental Analyses}\label{sup:ad_result} 

\subsection{\vlmfeat anchors}
\begin{figure}[h]
\centering
    \includegraphics[width=0.98\columnwidth]{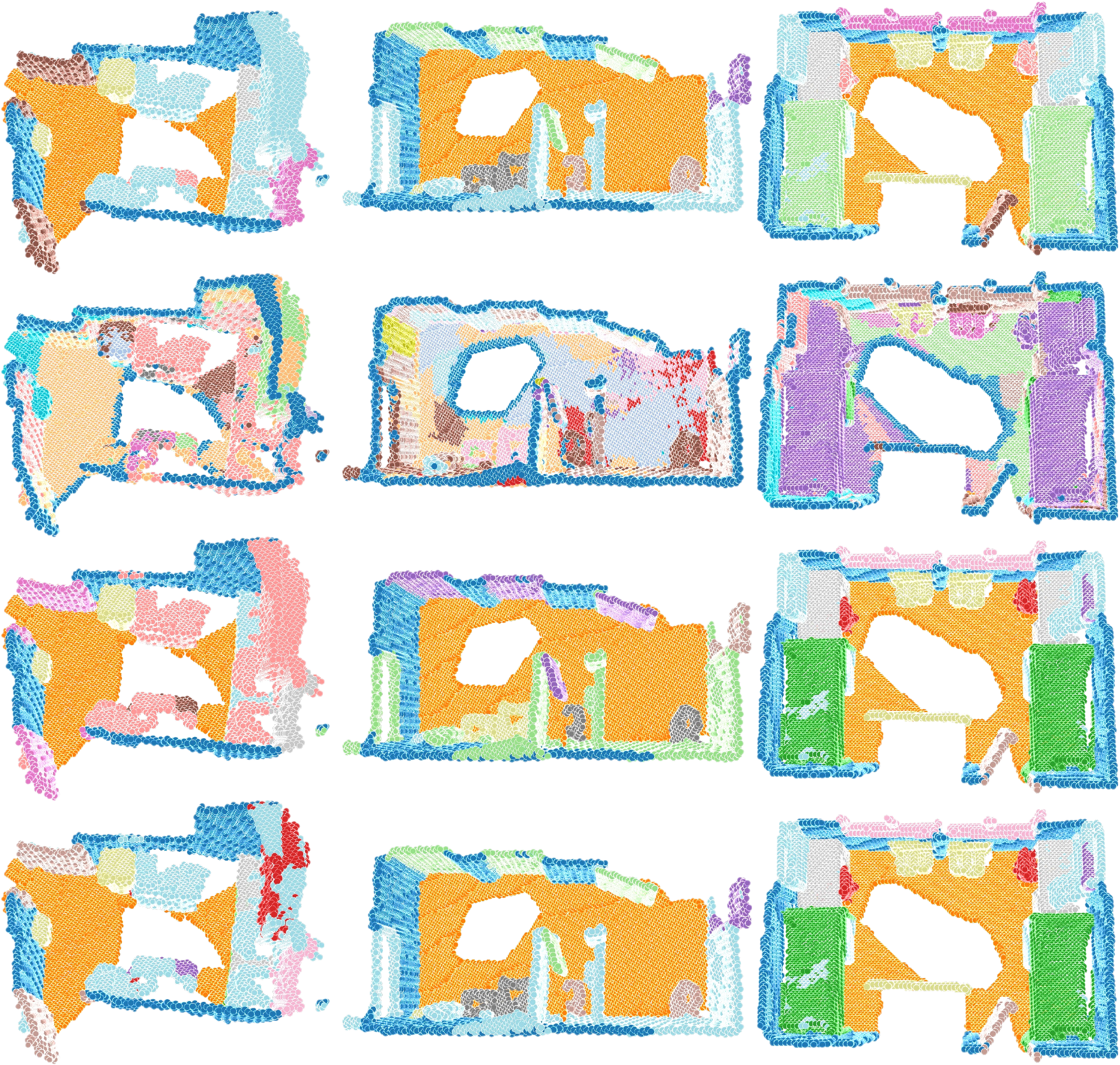}
    \put(0,195){\color{black}\footnotesize \rotatebox{90}{\textbf{GT}}}
    \put(0,120){\color{black}\footnotesize \rotatebox{90}{\textbf{OpenScene}}}
    \put(0,65){\color{black}\footnotesize \rotatebox{90}{\textbf{\ourmethod/wo}}}
    \put(0,15){\color{black}\footnotesize \rotatebox{90}{\textbf{\ourmethod}}}
    \caption{Zero-shot semantic segmentation results on ScanNet using OpenSeg feature extraction.
    (top row) ground-truth annotations, (second row) OpenScene (OpenSeg)~\cite{Peng2023}, (third row) \ourmethod/wo (OpenSeg) without using \vlmfeat anchors, and (bottom row) \ourmethod (OpenSeg). `wo' indicates the absence of \vlmfeat anchors.}
    \label{fig:anchor}
\end{figure}
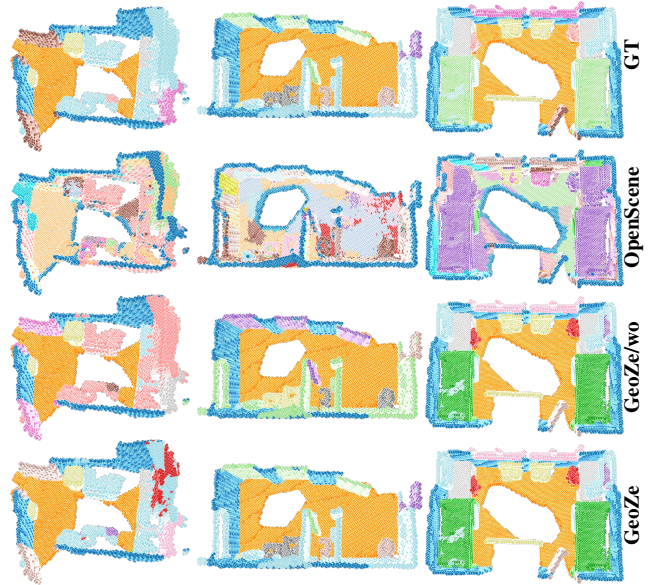

\vlmfeat anchor is designed to find important \vlmfeats that are more suitable for semantic alignment.
In our effort to cover a broad spectrum of categories and improve the anchors' general applicability, we conducte experiments using \vlmfeats derived from 32 point clouds.
This idea is inspired by the concept of a memory bank for constrastive learning.
We also produce \geofeat anchors, employing the same weight parameters as those used for calculating the \vlmfeat anchors. 
The purpose of these \geofeat anchors is to aid the anchor projection process, thereby mitigating issues of semantic misalignment.
This is achieved by searching for the closest \vlmfeat anchor for each point, taking into account both the similarities of \vlmfeats with \vlmfeat anchors and \geofeats with \geofeat anchors.
Fig.~\ref{fig:anchor} displays the results of zero-shot semantic segmentation on ScanNet~\cite{dai2017scannet}, using the OpenSeg as the feature extractor. This figure clearly shows the improved performance in zero-shot semantic segmentation achieved by incorporating \vlmfeat anchors (presented in the bottom row), especially when compared to the method that does not utilize \vlmfeat anchors, as seen in the third row of the figure.

\subsection{Classification visualization}
\begin{figure*}[t]
\centering
\subfloat[PointCLIPv2~\cite{zhu2023pointclip}]{\includegraphics[width=0.99\columnwidth]{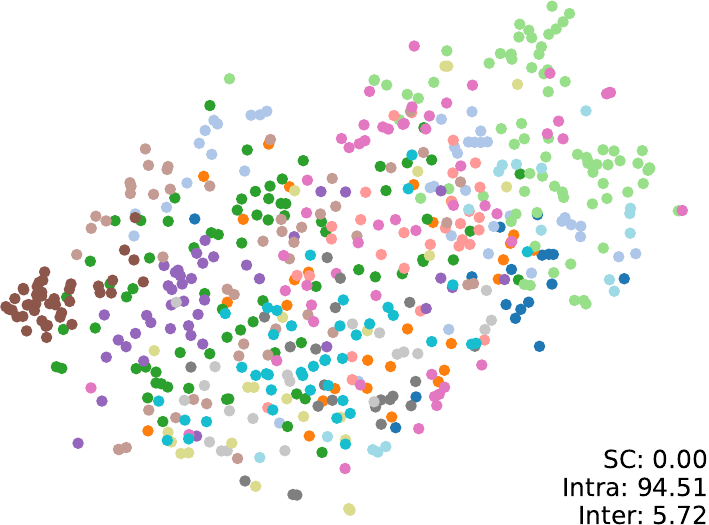}}
\hfil
\subfloat[\ourmethod]
{\includegraphics[width=0.99\columnwidth]{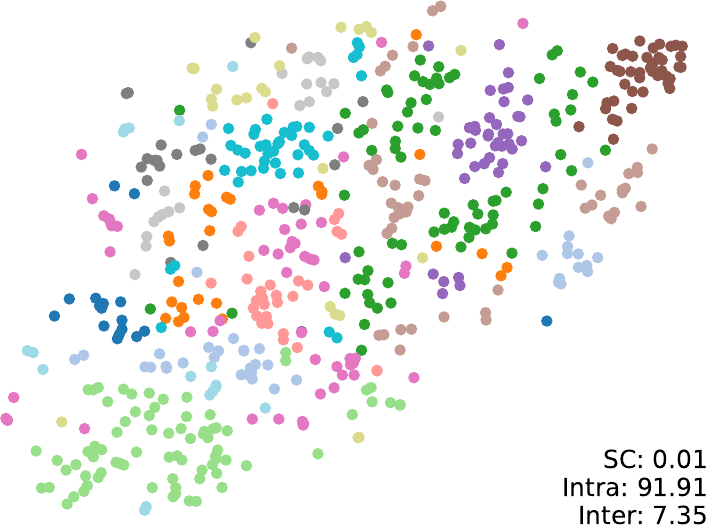}}
\caption{T-SNE embeddings of (a) PointCLIPv2~\cite{zhu2023pointclip} and (b) \ourmethod on ScanObjectNN~\cite{uy2019revisiting} (S-PB-T50-RS). 
\ourmethod produces better separated and grouped clusters for different categories, as evidenced by the superior silhouette coefficient (SC) and greater inter-cluster distance (inter), alongside a smaller intra-cluster distance (intra). 
}
\label{fig:suptsne}
\end{figure*}

In Fig.~\ref{fig:suptsne}, we report a t-SNE comparison between the class representations extracted with PointCLIPv2 and \ourmethod on ScanObjectNN~\cite{uy2019revisiting} (S-PB-T50-RS).
We quantify t-SNE clusters with three clustering metrics: Silhouette Coefficient, Inter-cluster Distance, and Intra-cluster Distance. 
From these metrics, we can confirm the efficacy of \ourmethod in better separating point features of diverse categories compared to PointCLIPv2.

\begin{figure}[t]
    \centering
    \includegraphics[width=1\linewidth]{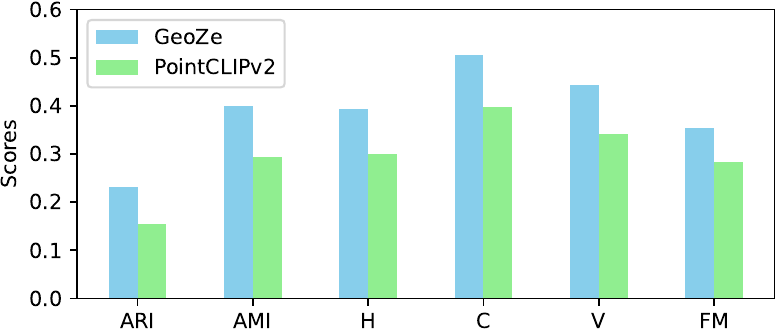}
    \caption{Comparison of clustering metrics (the higher the better) on ModelNet40 \cite{sharma2016vconv}.
    Metrics key: ARI: adjusted rand index, AMI: adjusted mutual information, H: homogeneity score, C: completeness score, V: V-measure,
FM: Fowlkes-Mallows score.}
\label{fig:sup_metric}
\end{figure}

Fig.~\ref{fig:sup_metric} compares the same global features using traditional clustering metrics for a more detailed assessment of the differences between \ourmethod and PointCLIPv2. 
For a comprehensive analysis of intra- and inter-cluster statistics, we consider six extrinsic clustering measures, that explicitly compare classification predictions with ground-truth annotations. The Adjusted Rand Index (ARI) evaluates the similarity of cluster assignments through pairwise comparisons. The Adjusted Mutual Information (AMI) assesses the agreement of cluster assignments. Homogeneity (H) gauges the proportion of instances from a single class in a cluster, akin to Precision. Completeness (C) measures the proportion of a given class's instances assigned to the same cluster, similar to Recall. 
The V-measure (V) quantifies clustering correctness using conditional entropy analysis. 
The Fowlkes-Mallows score (FM) evaluates clustering accuracy through the geometric mean of pairwise Precision and Recall. Higher scores in all these metrics indicate better performance. 
In Fig.~\ref{fig:sup_metric}, the histogram values are normalized, with the maximum value for each score set to 1. \ourmethod outperforms PointCLIPv2 in all metrics.

\subsection{Segmentation results on nuScenes}

\begin{figure*}[h]
\centering
    \includegraphics[width=1.0\linewidth]{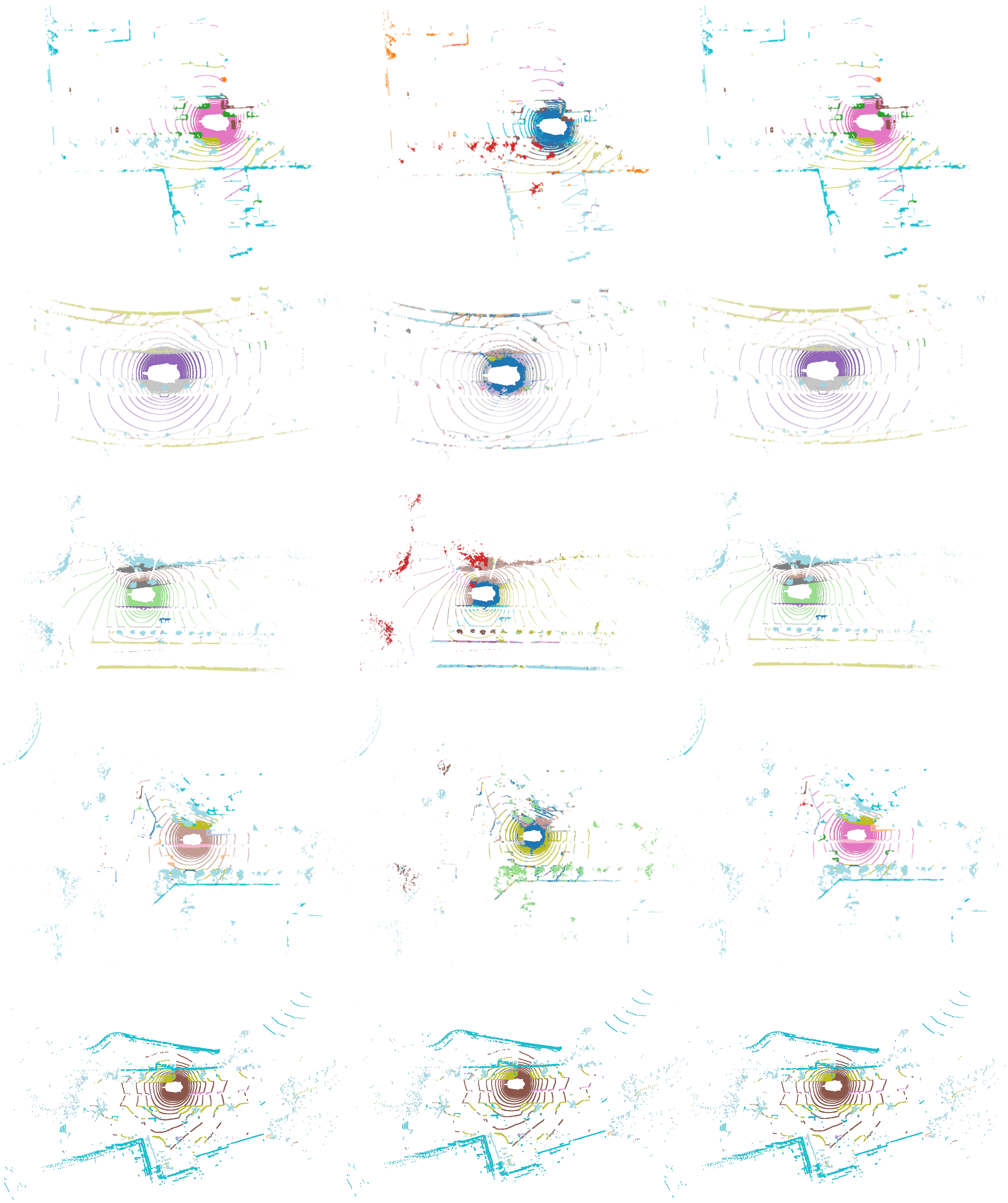}
    \put(-400,590){\color{black}\footnotesize{\textbf{GT}}}
    \put(-270,590){\color{black}\footnotesize{\textbf{OpenScene}}}
    \put(-80,590){\color{black}\footnotesize{\textbf{\ourmethod}}}
    \caption{Zero-shot semantic segmentation results on nuScenes~\cite{caesar2020nuscenes} using OpenSeg as feature extractor.
    (left column) ground-truth annotations, (middle column) OpenScene (OpenSeg)~\cite{Peng2023}, and (right column) \ourmethod.}
    \label{fig:nuScenes}
\end{figure*}

Fig.~\ref{fig:nuScenes} showcases a range of qualitative results obtained using our method, \ourmethod, on the nuScenes outdoor dataset~\cite{caesar2020nuscenes}. 
The effectiveness of \ourmethod is underscored by its ability to produce more semantically coherent segmented regions, a property primarily attributed to our novel clustering technique. Additionally, the synergy of local and global aggregation techniques bolsters the \vlmfeat, making it more geometrically aware. Another reason is the integration of \geofeat assignment, which plays a pivotal role in reducing semantic misalignment during anchor projection. The positive effects of our method are particularly visible at the boundaries of point clouds, where \ourmethod substantially lowers noise levels, outperforming the compared method OpenScene. This noise reduction is especially effective owing to the distinct geometric structures commonly found at these boundary zones. 

\subsection{\vlmfeat guided clustering}
\begin{figure*}[ht]
\centering
    \includegraphics[width=0.99\linewidth]{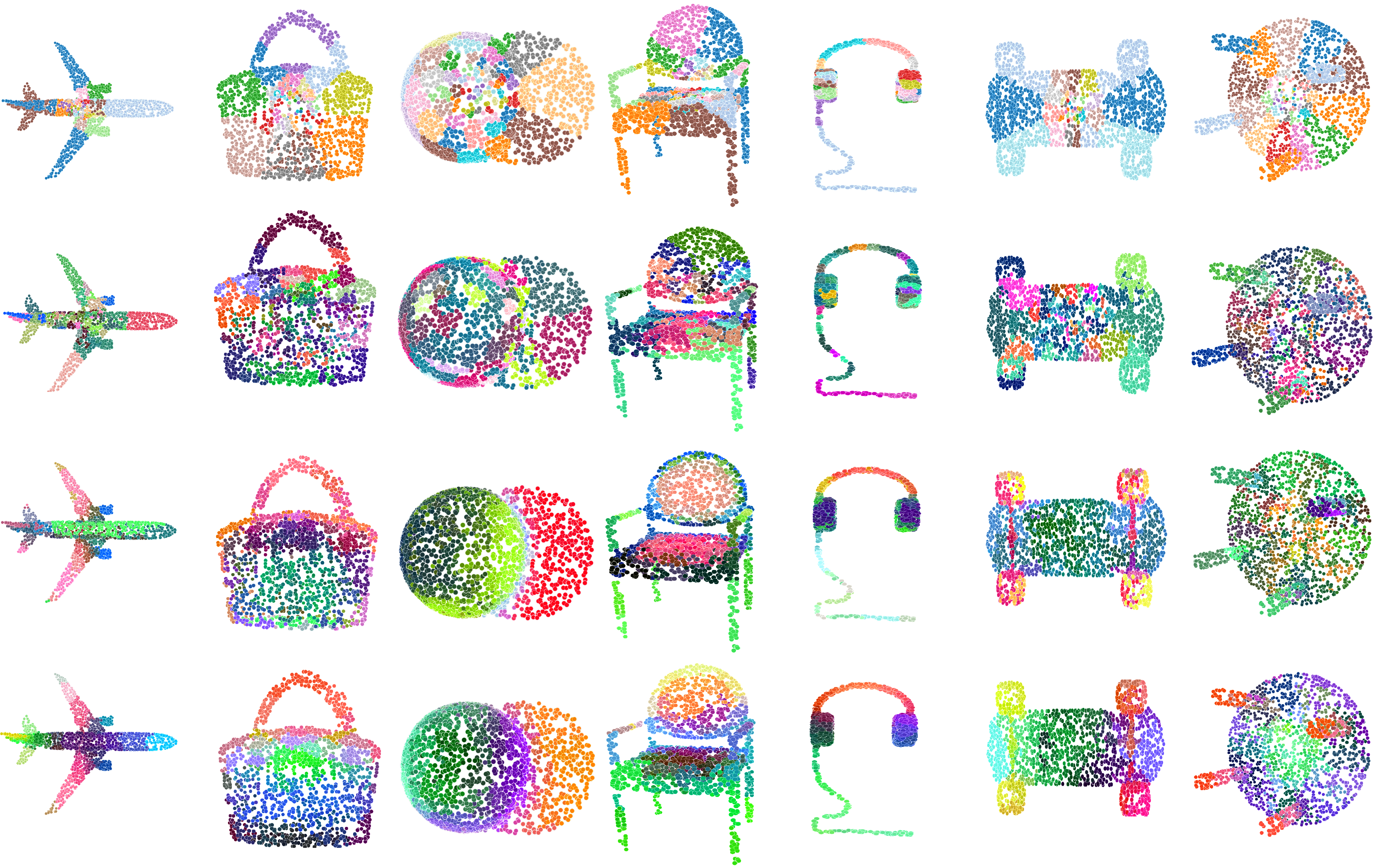}
    \put(-500,290){\color{black}\footnotesize \rotatebox{-90}{\textbf{Coord.}}}
    \put(-500,220){\color{black}\footnotesize \rotatebox{-90}{\textbf{Coord.+VLM}}}
    \put(-500,145){\color{black}\footnotesize \rotatebox{-90}{\textbf{Coord.+FPFH}}}
    \put(-500,55){\color{black}\footnotesize \rotatebox{-90}{\textbf{\ourmethod}}}
    \vspace{-1mm}
    \caption{Visualization of clustering results on ShapeNetPart~\cite{yi2016scalable} using various sources: 
    (top row) coordinates only;
    (second row) coordinates with \vlmfeats;
    (third row) coordinates and \geofeats (FPFH);
    (fourth row) coordinates and \geofeats (FPFH), guided by \vlmfeat similarity. Coord. represents Coordinates.}
    \label{fig:sup_clust}
\vspace{-3mm}
\end{figure*}

In this section, we demonstrate the enhancement in semantic clustering performance achieved by integrating geometric information into \vlmfeat. 
We use the clustering scores to compute each cluster's prototypical representation. Specifically, these prototypes are weighted averages of \vlmfeats, based on the clustering scores. Subsequently, each point is assigned to the prototype of its corresponding cluster. PCA projection is then applied to visualize the clustering results. 
Fig.~\ref{fig:sup_clust} showcases some qualitative clustering results on ShapeNetPart~\citep{yi2016scalable} using different point-level coordinates (Coord.) and representations. The top row demonstrates results using only coordinates for clustering, while the second row combines the coordinates with \vlmfeats. The third row integrates coordinates and \geofeats (FPFH), offering more meaningful partitioning compared to using just coordinates or coordinates with \vlmfeats. The bottom row features \ourmethod, which considers both coordinates and \geofeats (FPFH) for clustering, guided by \vlmfeat similarity, achieving the best clustering results (same parts tend to share colors).

\paragraph{Acknowledgement.} 
This work was supported by the PNRR project FAIR- Future AI Research (PE00000013), under the NRRP MUR program funded by the NextGenerationEU.

\end{document}